\documentclass[lettersize,journal]{IEEEtran}
\usepackage{amsmath,amsfonts}
\usepackage{algorithmic}
\usepackage{algorithm}
\usepackage{array}
\usepackage[caption=false,font=normalsize,labelfont=sf,textfont=sf]{subfig}
\usepackage{textcomp}
\usepackage{stfloats}
\usepackage{url}
\usepackage{verbatim}
\usepackage{graphicx}
\usepackage{cite}
\usepackage{times}
\usepackage{epsfig}
\usepackage{graphicx}
\usepackage{amsmath}
\usepackage{amssymb}
\usepackage{url}
\usepackage{array}
\usepackage{multirow}
\usepackage{adjustbox}
\usepackage{threeparttable}
\usepackage{algorithm}
\usepackage{setspace}
\usepackage[table,xcdraw]{xcolor}
\usepackage{amsmath}
\usepackage{amssymb}
\usepackage{multirow}
\usepackage{adjustbox}
\usepackage{threeparttable}
\usepackage{algorithmic}
\usepackage{utfsym}
\usepackage[algo2e,ruled,vlined]{algorithm2e}

 \usepackage[colorlinks]{hyperref}
 \hypersetup{
 	linkcolor=red,
 	filecolor=red,
 	citecolor=blue,      
 	urlcolor=magenta,
 }
\begin{document}

\title{MA-FSAR: Multimodal Adaptation of CLIP for Few-Shot Action Recognition}

\author{Jiazheng~Xing,  Chao~Xu, Mengmeng~Wang,
		  Guang Dai, Baigui Sun,  \\  Yong~Liu,~\IEEEmembership{Member,~IEEE},  Jingdong~Wang,~\IEEEmembership{Fellow,~IEEE}, and Jian~Zhao,~\IEEEmembership{Member,~IEEE}% <-this % stops a space
		\IEEEcompsocitemizethanks{\IEEEcompsocthanksitem Jiazheng~Xing and Yong Liu are with the Laboratory of Advanced Perception on Robotics and Intelligent Learning, College of Control Science and Engineering, Zhejiang University, Hangzhou 310027, Zhejiang, China. E-mail: jiazhengxing@zju.edu.cn, and yongliu@iipc.zju.edu.cn. Yong Liu and Jian Zhao are the corresponding authors.
 \IEEEcompsocthanksitem Chao Xu and Baigui Sun are with Alibaba Group. E-mail: xc264362@alibaba-inc.com and 
 sunbaigui85@gmail.com
 % \IEEEcompsocthanksitem Jun Dan is with the College of Information Science and Electronic Engineering, Zhejiang University, E-mail: danjun@zju.edu.cn
 \IEEEcompsocthanksitem Mengmeng Wang is with the College of Computer Science and Technology, Zhejiang University of Technology. E-mail: wangmengmeng@zjut.edu.cn.
  \IEEEcompsocthanksitem Guang Dai is with SGIT AI Lab,  State Grid Corporation of China. E-mail: guang.gdai@gmail.com.
  \IEEEcompsocthanksitem Jingdong Wang is with Baidu, E-mail: wangjingdong@baidu.com.% <-this % stops an unwanted space
 \IEEEcompsocthanksitem Jian Zhao is with EVOL Lab, Institute of AI (TeleAI), China, and School of Artificial Intelligence, Optics and Electronics (iOPEN), Northwestern Polytechnical University, Xi'an Shanxi, China. E-mail: zhaoj90@chinatelecom.cn.}
\thanks{Manuscript received April 19, 2021; revised August 16, 2021.}}

\markboth{Journal of \LaTeX\ Class Files,~Vol.~14, No.~8, August~2021}%
{Shell \MakeLowercase{\textit{et al.}}: A Sample Article Using IEEEtran.cls for IEEE Journals}


\maketitle

\begin{abstract}

Applying large-scale vision-language pre-trained models like CLIP to few-shot action recognition (FSAR)  can significantly enhance both performance and efficiency. While several studies have recognized this advantage, most of them resort to full-parameter fine-tuning to make CLIP's visual encoder adapt to the FSAR data, which not only costs high computations but also overlooks the potential of the visual encoder to engage in temporal modeling and focus on targeted semantics directly. To tackle these issues, we introduce MA-FSAR, a framework that employs the Parameter-Efficient Fine-Tuning (PEFT) technique to enhance the CLIP visual encoder in terms of action-related temporal and semantic representations. Our solution involves a Fine-grained Multimodal Adaptation, which is different from the previous attempts of PEFT in regular action recognition. Specifically, we first insert a Global Temporal Adaptation that only receives the class token to capture global motion cues efficiently. Then these outputs integrate with visual tokens to enhance local temporal dynamics by a Local Multimodal Adaptation, which incorporates text features unique to the FSAR support set branch to highlight fine-grained semantics related to actions. In addition to these token-level designs, we propose a prototype-level text-guided construction module to further enrich the temporal and semantic characteristics of video prototypes. Extensive experiments demonstrate our superior performance in various tasks using minor trainable parameters.

\end{abstract}

	\vspace{-15pt}

\section{Introduction}
% \IEEEPARstart{T}{his} file is intended to serve as a ``sample article file'' \cite{carreira2017quo}
\IEEEPARstart{F}{ew-shot} action recognition (FSAR) aims to quickly learn new action categories using limited labeled samples. Unlike conventional action recognition (AR)~\cite{li2023spatio, wang2022learning, wang2023actionclip, wang2016temporal, tong2022semi, wang2021multi, gao2020pairwise, xu2019semisupervised, chen2017deep}, FSAR is characterized by the extremely limited amount of labeled data available for each task and the wide variety of distinct task types. Therefore, FSAR necessitates the development of models capable of swiftly adapting to different tasks, making this endeavor exceptionally challenging. Previous approaches~\cite{cao2020few, zhu2018compound, zhang2020few, perrett2021temporal, thatipelli2022spatio, xing2023revisiting, wang2022hybrid, li2022hierarchical, huang2022compound, wang2023task, liu2022multidimensional, ji2023semantic} mainly focused on metric-based meta-learning paradigm and episode training to facilitate the transfer to new classes. However, relying solely on this paradigm still requires the model to spend much time training on different datasets, which somewhat hinders its application in the industry.

\begin{figure*} [t!]
	\centering
	\includegraphics[width=0.85\linewidth]{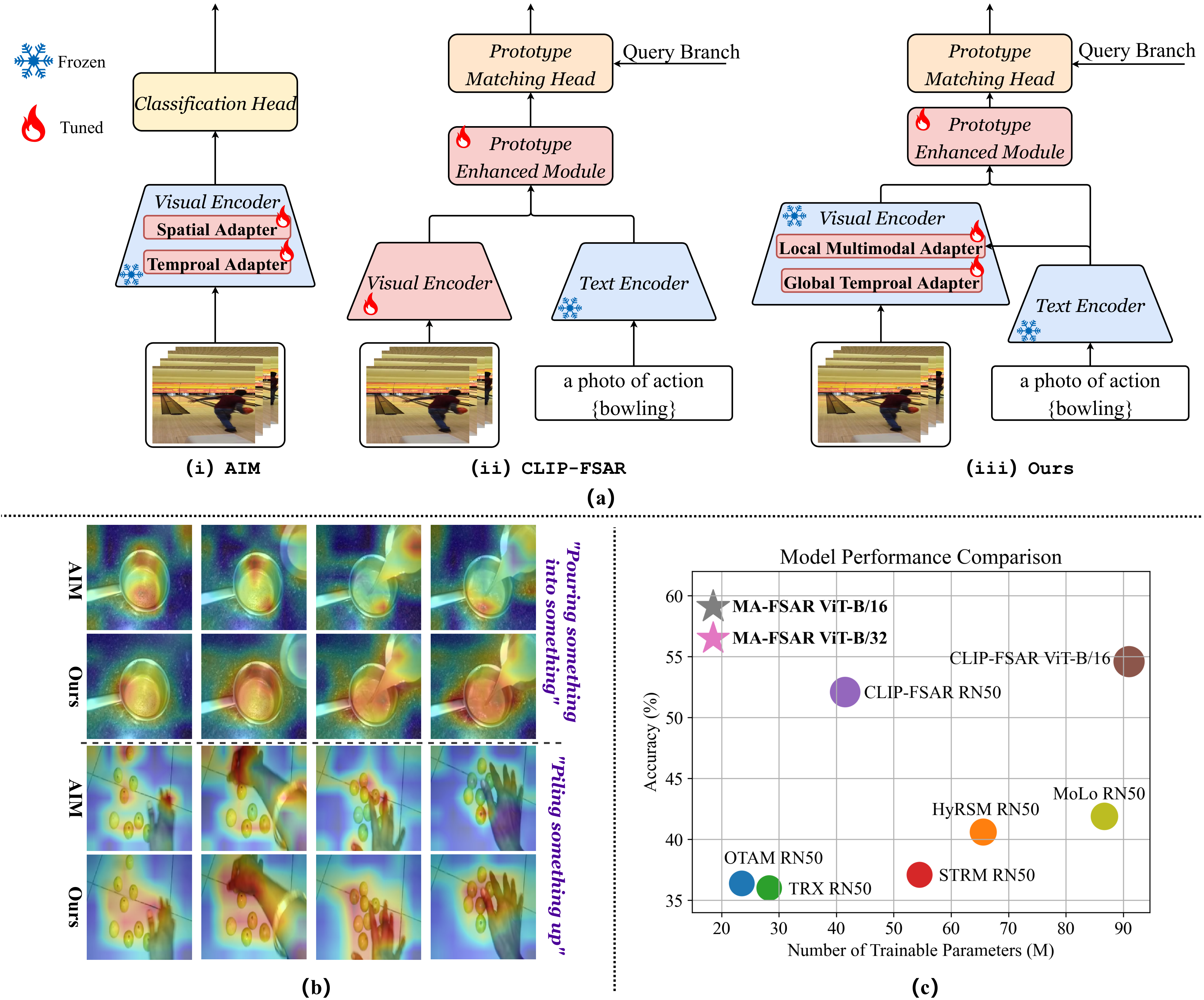}
 	\vspace{-10pt}

	\caption{(a): (i)  AIM~\cite{yang2023aim}, a method that successfully applied PEFT technology in action recognition; (ii) The support branch of CLIP-FSAR~\cite{wang2023clip}, a representative method that fully fine-tunes CLIP for few-shot action recognition; and (iii) the pipeline of our proposed method's support branch. (b): Visualization of the attention map at the visual encoder's last layer for the proposed MA-FSAR and AIM~\cite{yang2023aim}. AIM serves for action recognition as a classification task, whereas few-shot action recognition is a matching task. Therefore, for a fair comparison, both methods use the same few-shot temporal alignment metric, OTAM~\cite{cao2020few}. For the comparison result, the attention maps from our method are more focused on action-related objects due to the integration of text tokens and visual tokens in the visual encoder. (c):  Performance comparison of different few-shot action recognition methods in the SSv2-Small 5-way 1-shot task, including our \textbf{MA-FSAR}, OTAM~\cite{cao2020few}, TRX~\cite{perrett2021temporal}, STRM~\cite{thatipelli2022spatio}, HyRSM~\cite{wang2022hybrid}, MoLo~\cite{wang2023molo} and CLIP-FSAR~\cite{wang2023clip}. Bubble or star size indicates the recognition accuracy. Our \textbf{MA-FSAR} achieves the highest recognition accuracy with the least number of trainable parameters. }
	\label{fig:performance}
	\vspace{-5pt}
\end{figure*}

In recent years, more and more large-scale foundation vision-language models (VLM) have emerged, like CLIP~\cite{radford2021learning}, ALIGN~\cite{jia2021scaling}, and Florence~\cite{yuan2021florence}. As a consequence, researchers have actively delved into methods to effectively adapt these large models to their specific downstream tasks, such as action recognition~\cite{wang2023actionclip, ni2022expanding}, segmentation~\cite{rao2022denseclip,luddecke2022image,xu2022groupvit}, and object detection~\cite{gu2021open, zhao2022exploiting}. Undoubtedly, applying the ``pre-training, fine-tuning" paradigm leverages the power of robust pre-trained models, thus eliminating the need to train a network from scratch and obtain impressive performance.  
In few-shot action recognition, some methods such as CLIP-FSAR ~\cite{wang2023clip}, MVP-shot~\cite{qu2024mvp}, and CLIP-M${^2}$DF~\cite{guo2024multi} have made preliminary attempts, but they opt for full-parameter fine-tuning of the CLIP visual encoder with high computation costs. Such a substantial computational investment is merely to adapt the visual encoder to the FSAR data domain. Temporal modeling and action-related semantics distillation are left to be handled in subsequent prototype matching phases, as shown in Fig.~\ref{fig:performance}(a)(ii). In contrast, we argue that a better approach would be to utilize a \textit{minimal number of parameters to enable CLIP to directly mine motion cues and focus on dynamic target semantics}, with the prototype-level processing only as an additional supplementary module. 

Driven by these goals, we initially explore the possibility of tuning the visual encoder with a limited set of parameters, finding the Parameter-Efficient Fine-Tuning (PEFT) technique to be well-suited to our requirements. Its core idea is to keep the large pre-trained foundation model frozen and introduce trainable adapters ~\cite{yang2023aim,park2023dual, liu2023revisiting} or prompts ~\cite{wasim2023vita, ju2022prompting} for efficient fine-tuning to achieve robust performance among various tasks. Although PEFT has been experimented with in action recognition (AR), like AIM~\cite{yang2023aim}, Vita-CLIP~\cite{wasim2023vita}, and ST-Adapter~\cite{pan2022st}, its application in FSAR is yet to be explored. Since FSAR is a matching task rather than the classification task as AR, more discriminative features are required to establish class prototype centers for each task, and thus it is not appropriate to directly borrow the adapter leveraged in AR. Concretely, the reasons are two folds, 1) as shown in Fig.~\ref{fig:performance}(a)(i), they employ a temporal adapter to integrate temporal information into the CLIP encoder, but the visual tokens encompass numerous cues unrelated to the action, which can diminish the discriminativeness of temporal features. 2) The support set in FSAR contains distinctive labeled text information that can be used as textual features to bolster the semantic discriminativeness of dynamic features.

% Additionally, we include a Joint Adaptation to tune the final discriminative representations. 
To address these issues, we propose a novel method, dubbed \textbf{MA-FSAR}, a shot for \textbf{M}ultimodal \textbf{A}daptation of CLIP for \textbf{F}ew-\textbf{S}hot \textbf{A}ction \textbf{R}ecognition. Our solution is built upon PEFT and incorporates a Fine-grained Multimodal Adaptation (FgMA) tailored for FSAR without altering the original CLIP weights, thus enabling action-related temporal awareness and semantic mining capabilities to be seamlessly integrated into the CLIP visual encoder in an effective and efficient manner. Specifically, we first introduce a Global Temporal Adaptation that processes only the class token that contains the precise semantics of each frame to efficiently capture global dynamic cues. These outputs provide an accurate motion prior to the subsequent Local Multimodal Adaptation to guide the local visual tokens in learning spatiotemporal details. Notably, this module is also capable of integrating text features that are specific to the FSAR support set, thereby highlighting the fine-grained semantics associated with actions. As shown in Fig.~\ref{fig:performance}(b), compared to the adapter used in AIM~\cite{yang2023aim}, which does not incorporate textual features, our method's attention maps are more focused on action-related objects. 
After completing the token-level designs of CLIP, we propose a Text-guided Prototype-level Construction Module (TPCM) to further enrich the temporal and semantic characteristics of video prototypes. In summary, our proposed plug-and-play MA-FSAR can flexibly integrate with any common FSAR matching metric, ensuring the efficient and effective application of CLIP to FSAR. Extensive experiments unequivocally demonstrate that our method attains exceptional performance while employing the fewest tunable parameters, as shown in Fig.~\ref{fig:performance}(c). We make the following contributions:

\begin{itemize}
\item  We propose a novel method, \textbf{MA-FSAR}, to refine CLIP's visual encoder at the token level for FSAR by introducing the Fine-grained Multimodal Adaptation with the enhancement of the action-related temporal and semantic representations, which is fast, efficient, and cost-effective in training.
\item At the prototype level, we propose a Text-guided Prototype Construction Module to further enhance the temporal and semantic representation of video prototypes.
\item  Experiments demonstrate that our method performs excellently using any metric in various task settings on five widely used datasets with minimal trainable parameters.

% Our plug-and-play method can be used in any different few-shot action recognition temporal alignment metric.
% \item Extensive experiments on five widely used datasets have shown that our method can achieve outstanding performance with minor trainable parameters.
\end{itemize}
\vspace{-8pt}
\section{Related Works}
\subsection{Few-shot Learning}
Research on few-shot learning can be mainly classified into adaptation-based and metric-based methods. The former aims to find a network initialization that can be fine-tuned for unknown tasks using limited labeled data, called \textit{gradient by gradient}. Classical adaptation-based approaches include MAML~\cite{finn2017model} and Reptile~\cite{nichol2018reptile}, with further in-depth research found in ~\cite{li2017meta, wang2020m}. The latter aims to acquire knowledge of feature space and compare task features using various matching strategies, referred to as \textit{learning to compare}. Representative methods include Prototypical Networks~\cite{snell2017prototypical} and Matching Networks~\cite{vinyals2016matching}, with numerous approaches~\cite{yoon2019tapnet, ye2020few, doersch2020crosstransformers, li2019finding} aiming to make improvements based on these models.
% Despite the potential benefits of adaptation-based methods (\textit{e.g.}, MetaUVFS~\cite{patravali2021unsupervised}), these approaches have received limited attention in few-shot action recognition due to their high computational requirements and extensive experimental time.
\subsection{Few-shot Action Recognition}
The core concept of few-shot action recognition (FSAR) is similar to few-shot learning (FSL), but the inclusion of the temporal dimension increases the complexity of the problem. Adaptation-based methods such as MetaUVFS~\cite{patravali2021unsupervised} have received limited attention in FSAR due to their high computational demands and extensive experimental time. Therefore, existing research predominantly emphasizes metric-based learning approaches with varying focuses. On the one hand, some methods focus on class prototype matching metric strategies.  OTAM~\cite{cao2020few} introduces a temporal alignment metric to calculate the distance value between query and support set videos. TRX~\cite{perrett2021temporal} matches each query sub-sequence with all sub-sequences in the support set, facilitating correspondences between different videos. And, HyRSM~\cite{wang2022hybrid} proposes a bidirectional Mean Hausdorff Metric that exhibits robustness to complex actions. On the other hand, certain approaches aim to enhance feature or class prototype representations. STRM~\cite{thatipelli2022spatio} adopts local and global enrichment modules for features' spatiotemporal modeling. HyRSM~\cite{wang2022hybrid} utilizes hybrid relation modeling to learn task-specific embeddings. And, TADRNet~\cite{wang2023task} simultaneously considers both global and local characteristics and proposes a hybrid semantic attention module for enhancing the discriminability of samples. Recently, with the development of large foundation vision-language models, their application in FSAR is receiving increasing attention. The most representative work, CLIP-FSAR~\cite{wang2023clip}, refines CLIP for FSAR by fully fine-tuning CLIP's visual encoder and designing a prototype modulation module to enhance multimodal representations at the prototype level.  However, despite its significant computational overhead, it falls short in temporal modeling and multimodal feature fusion within the CLIP visual encoder.
% Another approach is AMeFu-Net~\cite{fu2020depth}, which exploits depth information to assist in learning.

\subsection{Parameter-efficient Fine-tuning (PEFT) for Vision Models}
 Parameter-efficient Fine-tuning (PEFT) technique, initially employed in Natural Language Processing ~\cite{houlsby2019parameter, lester2021power, zaken2021bitfit, hu2021lora}, has exhibited impressive advancements in Computer Vision in recent times. Its application in video understanding can be broadly categorized into two main approaches: Adapter-based and Prompt-tuning-based. The design of the Adapter originates from~\cite{houlsby2019parameter}. It adds two fully connected layers (FC) with residual structures in each transformer layer to fine-tune the model, where the original transformer is frozen and only the adapter layer is trained during the process. Inspired by this, AIM~\cite{yang2023aim} applies the Adapter technique in action recognition. In each Vision Transformer (ViT)~\cite{dosovitskiy2020image} block, AIM designs three adapters for spatial, temporal, and joint adaptation, achieving excellent results. Besides that, ST-Adapter~\cite{pan2022st} introduced a parameter-efficient spatiotemporal adapter, effectively leveraging the capabilities of CLIP's image models for video understanding. As for Prompt-tuning, it refers to the flexible adjustment of prompts, which can significantly impact the final performance of the model. The pioneering use of Prompt-tuning in the visual domain is by VPT~\cite{jia2022visual}. It introduces learnable prompts within ViT while freezing the other training parameters in the network and achieves impressive results in image-related downstream tasks. Inspired by this, Vita-CLIP~\cite{wasim2023vita} designs the Prompt-tuning method specifically for videos, which proposes the learnable video summary tokens, frame-level prompts, and video-level prompts, achieving impressive results. In this work, we incorporate Parameter-Efficient Fine-Tuning (PEFT) techniques into the FSAR task to enhance the performance of the CLIP visual encoder without demanding excessive computational resources.
 
 % Due to Adapter's simplicity and AIM's success in action recognition, we choose the Adapter-based method as our PEFT method.

\section{Method}
\subsection{Problem Formulation}
\label{problem_formu}
In few-shot action recognition, the objective is to classify an unlabeled query video into one of the $M$ action categories within the support set, with only  $K$ limited samples per action class. This is considered an  $M$-way $K$-shot task.  Similar to previous researches~\cite{cao2020few, zhu2018compound, zhang2020few, perrett2021temporal, thatipelli2022spatio, xing2023revisiting, wang2022hybrid, li2022hierarchical, huang2022compound}, we follow the episode training framework, where episodes are randomly selected from a vast pool of collected data. In each episode, we assume that the set $\mathcal{S}$ comprises $M\times K$
samples originating from  $M$ different action classes. Specifically, 
$S_k^m=\left\{s_{k1}^m,s_{k2}^m,\cdots,s_{kT}^m\right\}$ denotes the $k$-th video in class  $m \in\left\{1,\cdots,M\right\}$, randomly sampled with $T$ frames. And, the query video is represented as $ Q=\left\{q_1,q_2,\cdots,q_T\right\}$, also sampled with $T$ frames.
\vspace{-5pt}
% CLIP can simultaneously encode input images and texts and map them into the same vector space. It can perform cross-modal reasoning and achieve mutual conversion between images and texts.
 % CLIP has been pre-trained on 400 million web-crawled image-text pairs, making the model highly generalizable. 
 % the limited labeled video samples in each task can be significantly enhanced by extracting semantic information from label texts and associating it with the corresponding video features.
% Leveraging the multimodal advantages of CLIP, the labeled texts of support set videos can be used to enhance the semantic discriminativeness of corresponding video features in few-shot action recognition (FSAR). 
 
\subsection{Architecture Overview}
In this work, we choose CLIP~\cite{radford2021learning} as the pre-trained foundation vision-language model, which features a dual-encoder structure consisting of visual and text encoders. It can perform cross-modal reasoning and achieve mutual conversion between images and texts. For the vision branch, we select the ViT (Vision Transformer) ~\cite{dosovitskiy2020image} architecture from CLIP as our visual encoder due to its powerful feature encoding capabilities and its flexible token-based learning network structure,  which facilitates the application of the PEFT technique. 
For the text branch, labeled input texts are typically combined with prompt templates before passing through the encoder, and the method for selecting these templates is detailed in Sec.~\ref{NA}).

\begin{figure*} [ht!]
	\centering
	\includegraphics[width=0.95\linewidth]{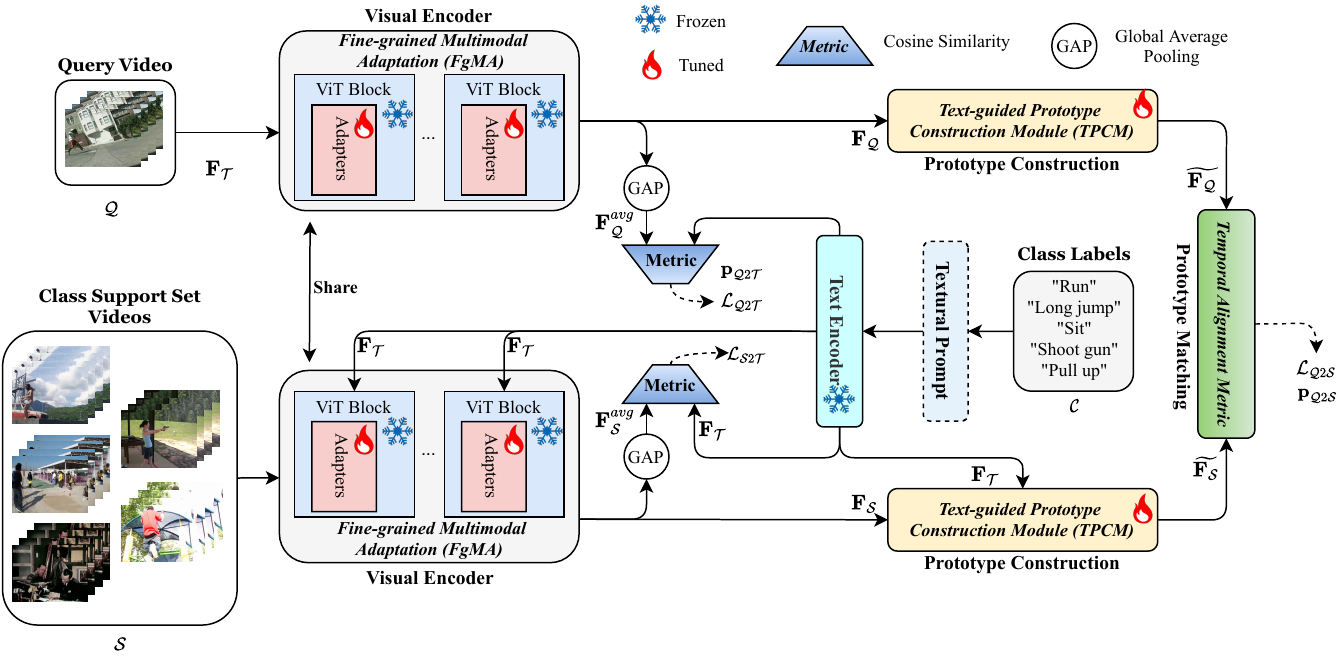}
 \vspace{-16pt}
	\caption{Overview of \textbf{MA-FSAR}. For simplicity and convenience, we focus on a specific scenario: the 5-way 1-shot task with a query set $\mathcal{Q}$ containing a single video. The support set video features $\textbf{F}_{\mathcal{S}}$ and query video feature $\textbf{F}_{\mathcal{Q}}$ are obtained by the visual encoder with the Fine-grained Multimodal Adaptation (FgMA). Text features $\textbf{F}_{\mathcal{T}}$ are obtained through a text encoder. The Text-guided Prototype Construction Module (TPCM)  generates the final features before the prototype matching, denoted as $\widetilde{\textbf{F}_\mathcal{S}}$ and $\widetilde{\textbf{F}_\mathcal{Q}}$. The probability distribution $\textbf{p}_{\mathcal{Q}2\mathcal{T}}$ is obtained using cosine similarity metric, and $\textbf{p}_{\mathcal{Q}2\mathcal{S}}$ is calculated using prototype matching metric. The loss $\mathcal{L}_{\mathcal{Q}2\mathcal{S}}$ is the standard Cross-Entropy loss, while  $\mathcal{L}_{\mathcal{S}2\mathcal{T}}$ and $\mathcal{L}_{\mathcal{Q}2\mathcal{T}}$ are Kullback-Leibler divergence (KL) losses.}
	\label{fig:pipeline}
			\vspace{-12pt}
\end{figure*}

Our overall architecture is illustrated in Fig.~\ref{fig:pipeline}. For the frame-selecting strategy, we use the approach from TSN~\cite{wang2016temporal}, which divides the input video sequence into $T$ segments and extracts snippets from each segment. For simplicity and convenience, we will focus on a specific scenario: the 5-way 1-shot problem with a query set $\mathcal{Q}$ containing a single video. In this pipeline, the query video $Q=\left\{q_1,q_2,\cdots,q_T\right\}$ and the class support set videos $S^m=\left\{s_{1}^m,s_{2}^m,\cdots,s_{T}^m\right\}$ \big($S^m\in\mathcal{S}=\left\{S^1, S^2,\cdots, S^5\right\}$\big) pass through the visual encoder with the Fine-grained Multimodal Adaptation (FgMA) to obtain the query feature $\textbf{F}_\mathcal{Q}$ and the support features $\textbf{F}_{\mathcal{S}}^m$($\textbf{F}_{\mathcal{S}}^m\in\textbf{F}_\mathcal{S}$) in each episode. And, the text label descriptions combined with the prompt template $C^m \big(C^m \in \mathcal{C}=\left\{C^1, C^2,\cdots, C^5\right\}\big)$ pass through the text encoder to obtain text features $\textbf{F}_{\mathcal{T}}^m$\big($\textbf{F}_{\mathcal{T}}^m \in \textbf{F}_{\mathcal{T}}\big)$. Then we apply global average pooling operation to $\textbf{F}_{\mathcal{S}}$ and  $\textbf{F}_{\mathcal{Q}}$ to obtain ${\textbf{F}_{\mathcal{S}}^{avg}}$ and ${\textbf{F}_{\mathcal{Q}}^{avg}}$. The Kullback-Leibler divergence losses $\mathcal{L}_{\mathcal{S}2\mathcal{T}}$ and $\mathcal{L}_{\mathcal{Q}2\mathcal{T}}$ are calculated by the cosine similarity metric between ${\textbf{F}_{\mathcal{S}}^{avg}}$, ${\textbf{F}_{\mathcal{Q}}^{avg}}
$, and $\textbf{F}_{\mathcal{T}}$, facilitating the adaptation of CLIP to FSAR domain. And, the probability distribution $\textbf{p}_{\mathcal{Q}2\mathcal{T}}$ is derived using the cosine similarity metric. Then, $\textbf{F}_{\mathcal{S}}$ and  $\textbf{F}_{\mathcal{Q}}$ are passed through the Text-guided Prototype Construction Module (TPCM) with weight sharing to obtain the final features before the prototype matching, denoted as $\widetilde{\textbf{F}_\mathcal{S}}$ and $\widetilde{\textbf{F}_\mathcal{Q}}$. Finally, the enhanced features are fed into the prototype matching metric to obtain the probability distribution $\textbf{p}_{\mathcal{Q}2\mathcal{S}}$ and loss  $\mathcal{L}_{\mathcal{Q}2\mathcal{S}}$.

	\vspace{-10pt}

\subsection{Fine-grained Multimodal Adaptation (FgMA)}
\label{FgMA_module}
% and enhance temporal modeling and targeted semantics focusing at the token level
% To minimize trainable parameters and avoid overfitting while leveraging video spatiotemporal and text semantic information, we refine CLIP for few-shot action recognition with lightweight adapters. These adapters combine bi-modal information for temporal and multimodal modeling, each playing a role in different relationship learning within the network.
We employ the Parameter-Efficient Fine-Tuning (PEFT) technique to refine CLIP for the few-shot action recognition (FSAR) domain with minimal trainable parameters.  Due to Adapter's~\cite{houlsby2019parameter} simplicity and adapter-based method's~\cite{yang2023aim} success in action recognition, we propose the Fine-grained Multimodal Adaptation (FgMA) tailored for FSAR, which can enhance temporal modeling and targeted semantics focusing at the token level. This approach freezes the pre-trained image and text encoders during training while introducing new, lightweight, learnable adapters. 
 
In this subsection, since we explore how to design adaptations for the Vision Transformer in FSAR, we first provide an overview of the conventional ViT Block. Consider a video clip $V\in \mathbb{R} ^{T\times H\times W\times 3}
$, where $H,W$ represent the spatial size and $T$ represents the number of frames. Each frame $t\in\left\{ 1\cdots T \right\} $ is divided into $N$ non-overlapping square patches $\left\{ \textbf{x}_{t,i} \right\} _{i=1}^{N}\in \mathbb{R} ^{P^2\times 3}$ of size $P\times P$, with the total number of patches being $N = HW/P^2$. Then the patches $\left\{ \textbf{x}_{t,i} \right\} _{i=1}^{N}\in \mathbb{R} ^{P^2\times 3}$ are then projected into the patch embeddings $\textbf{x}_{t,p}\in \mathbb{R} ^{N \times D}$ through a linear projection $\textbf{E}\in\mathbb{R} ^{3P^2 \times D}$. An additional learnable [class] token $\textbf{x}_{cls} \in \mathbb{R} ^{D}$  is added to the embedded patch sequence $\textbf{x}_{t,p}$ for each frame, resulting in  $\textbf{x}_{t}^{(0)} = \left[ \textbf{x}_{cls}; \textbf{x}_{t,p} \right] \in\mathbb{R} ^{(N+1) \times D}$. The final per-frame token sequence fed into the ViT blocks is given by:
    \begin{equation}\label{}
    \textbf{z}_{t}^{(0)} = \textbf{x}_{t}^{(0)} + \textbf{e}_{pos}
    \end{equation}
where $\textbf{e}_{pos} \in \mathbb{R} ^{(N+1) \times D}$  represents the spatial position encoding. As shown in Fig.~\ref{fig:FgMA}(b), each ViT block comprises several components: a multiheaded self-attention (MSA) mechanism, a multilayer perceptron (MLP) layer, layer normalization (LN), and skip connections. Formally, the computation of a ViT block can be formulated as: 
    \begin{equation}\label{}
    {\textbf{z}'}_{t}^{(l)}=\textbf{z}_{t}^{(l-1)}+\mathrm{MSA}\left( \mathrm{LN}\left( \textbf{z}_{t}^{(l-1)} \right) \right)
    \end{equation}
    \begin{equation}\label{}
    {\textbf{z}}_{t}^{(l)}={\textbf{z}'}_{t}^{(l)}+\mathrm{MLP}\left( \mathrm{LN}\left({\textbf{z}'}_{t}^{(l)} \right) \right)
    \end{equation}
where $\textbf{z}_{t}^{(l-1)}$ and $\textbf{z}_{t}^{(l)}$ represent per-frame input and the output of the $l$-th ViT block, respectively.  And the video level representation at the $l$-th layer can be represented as $\textbf{z}^{(l)} = \left[\textbf{z}^{(l)}_0 \cdots \textbf{z}^{(l)}_t \cdots \textbf{z}^{(l)}_T \right]$. 
% As shown in Fig.~\ref{fig:FgMA}(a), Adapter has a straightforward structure that includes two fully connected layers (FC), an activation layer, and a residual connection. The first FC layer maps the input to a lower dimension, while the second FC layer maps the input back to its original dimension. The support and query set branches' network structures are represented in Fig.~\ref{fig:FgMA}(c) and Fig.~\ref{fig:FgMA}(d), respectively.Therefore, by leveraging the textual category information of the support videos, the semantic representation of class prototypes in few-shot tasks can be enhanced. Moreover, we reuse the pre-trained self-attention layer in the image model for temporal and multimodal adaptation to minimize the number of trainable parameters. By changing the dimensions of the input, the self-attention layer can be used in different ways. In what follows, we introduce three types of adaptation.

As for our Fine-grained Multimodal Adaptation (FgMA), it can be divided into three parts: Global Temporal Adaptation, Local Spatiotemporal/Multimodal Adaptation, and Joint Adaptation. Each Adaptation contains a frozen attention layer and a trainable adapter with a straightforward structure that includes two fully connected layers (FC), an activation layer, and a residual connection, as depicted in  Fig.~\ref{fig:FgMA}(a).  The key to the aforementioned two Adaptation lies in controlling the attention layer to perform feature modeling with different tendencies across various dimensions at the token level through fine-tuning the Adapter. In FSAR, the label information of the support set is known, while that of the query set is unknown, resulting in differing network structures for each, shown in Fig.~\ref{fig:FgMA} (c) and (d). In what follows, we introduce three types of Adaptation:

% \vspace{-4mm}
% Since videos have an additional temporal dimension compared to images, temporal modeling is crucial for video tasks. Based on this, we design temporal adaptation for temporal modeling. To reduce computational costs, we only use the [class] token $\textbf{x}_{cls}$ as the input for temporal modeling.
\subsubsection{\textbf{Global Temporal Adaptation}}
\begin{figure} [t!]
		\centering
	\includegraphics[width=\linewidth,height=1.55\linewidth]{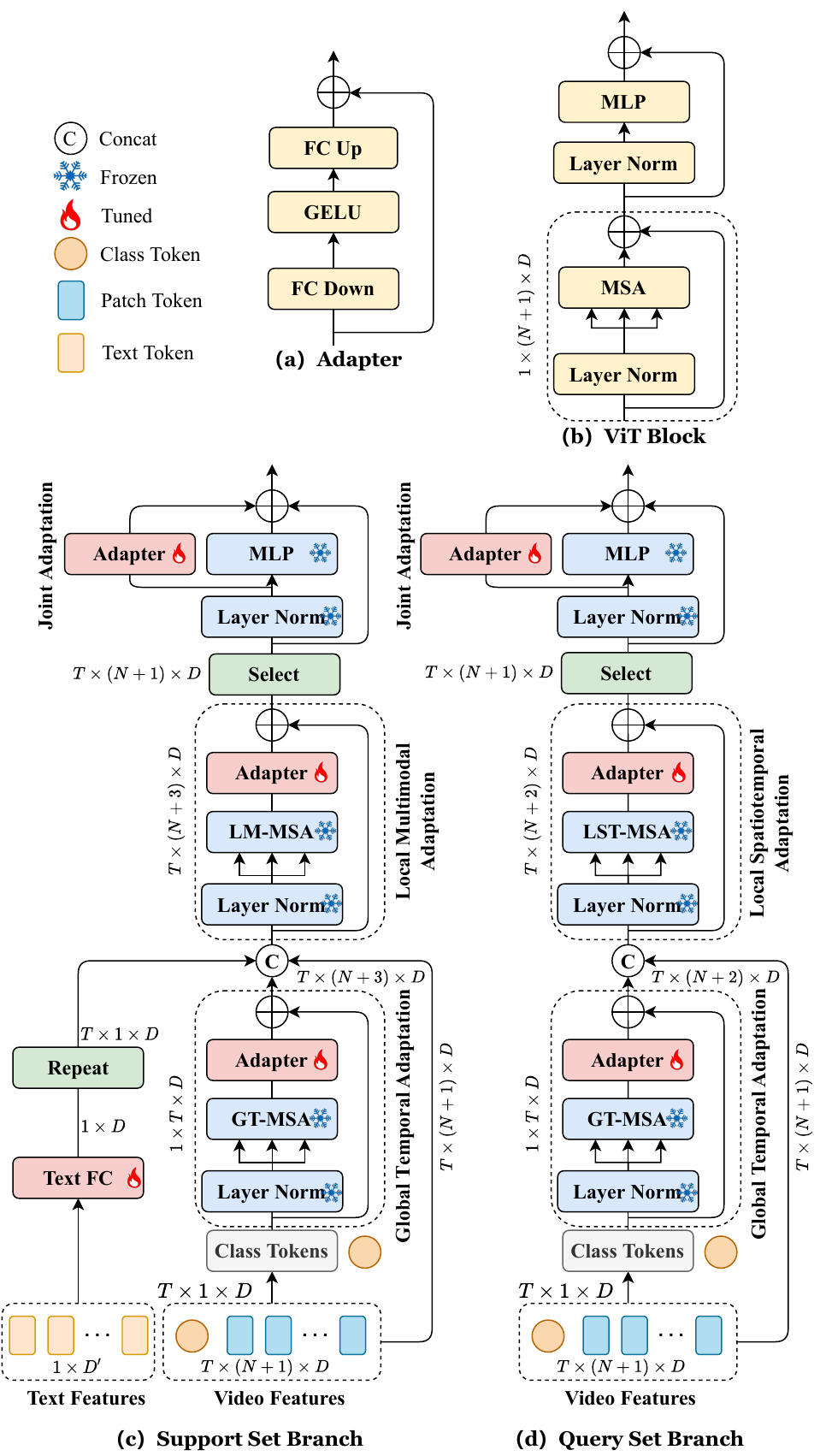}
 \vspace{-14pt}
		\caption{(a) shows the structure of the Adapter~\cite{houlsby2019parameter}, and (b) shows the structure of a standard ViT~\cite{dosovitskiy2020image} block. (c) and (d) illustrate the fine-grained multimodal adaptation of each ViT block for the support and query set branch. Note that GT-MSA, LM-MSA, and LST-MSA share weights but are applied to different inputs with different motivations for global temporal, local multimodal, and local spatiotemporal modeling.}
		\label{fig:FgMA}
  \vspace{-14pt}
	\end{figure}

 Temporal modeling is essential for FSAR, but directly involving all visual tokens in exploring temporal relationships would lead to significant computational overhead. Meanwhile, the visual tokens encompass numerous cues unrelated to the action, and applying all of the tokens might affect the salience of capturing temporal signals. Therefore, we introduce the Global Temporal Adaptation, which only processes the class token extracted from its visual tokens representing the accurate semantics of each frame, capturing global motion cues while reducing computational costs.  
 Specifically, for the $l^{th}$ layer, given the input video [class] token embedding $\textbf{x}_{cls}^{(l-1)}\in \mathbb{R} ^{T \times 1 \times D}$, we reshape it into $\textbf{x}_{TA}^{(l-1)}\in \mathbb{R} ^{1 \times T \times D}$. Then we feed $\textbf{x}_{TA}^{(l-1)}$ into Global Temporal Adaptation to capture the global dynamic cues between multiple frames, given by:
\begin{equation}\label{}
    {\textbf{x}}_{TA}^{(l)}={\textbf{x}}_{TA}^{(l-1)}+\mathrm{Adapter}\left( \mathrm{GT\text{-}MSA}\left(\mathrm{LN}\left({\textbf{x}}_{TA}^{(l-1)} \right) \right) \right)
\end{equation}
where $ {\textbf{x}}_{TA}^{(l-1)}$ and  $ {\textbf{x}}_{TA}^{(l)}$ denote the 
Global Temporal Adaptation input and output of the $l^{th}$ transformer block. Self-attention $\mathrm{GT\text{-}MSA}$ operates on the temporal dimension $T$ to learn the global temporal relationships between multiple frames. The Adapter structure maintains the same configuration as illustrated in Fig.~\ref{fig:FgMA}(a). However, the skip connection is removed to separate the influence of the adaptation during the initial training phase.

 % After the Global Temporal Adaptation, we aim to integrate spatiotemporal information with text semantic information to perform multimodal adaptation. 

\subsubsection{\textbf{Local Spatiotemporal/Multimodal Adaptation}} \label{MA}
The outputs of the Global Temporal Adaptation provide accurate global temporal priors. Our next step is to inject these global temporal cues into the visual tokens of each frame and introduce the Local Spatiotemporal Adaptation to guide local visual tokens in performing local spatiotemporal modeling. Specifically, in the query branch, we can concatenate each frame's input visual tokens $\textbf{z}^{(l-1)}$ and the corresponding global temporal token $\textbf{x}_{TA}^{(l)}$ along the spatial dimension to obtain ${\textbf{z}}_{STA\text{-}Q}^{(l-1)}=\left[ \textbf{z}^{(l-1)}; \textbf{x}_{TA}^{(l)} \right] \in \mathbb{R}^{T \times (N+2) \times D}$. Then we feed ${\textbf{z}}_{MA-Q}^{(l-1)}$ into the Local Spatiotemporal Adaptation as shown in Fig.~\ref{fig:FgMA}(d), written by: 
 \begin{equation}\label{}
    {\textbf{z}}_{STA\text{-}Q}^{(l)}={\textbf{z}}_{STA\text{-}Q}^{(l-1)}+\mathrm{Adapter}\left( \mathrm{LST\text{-}MSA}\left(\mathrm{LN}\left({\textbf{z}}_{STA\text{-}Q}^{(l-1)} \right) \right) \right)
\end{equation}
 where ${\textbf{z}}_{STA\text{-}Q}^{(l-1)}$ and ${\textbf{z}}_{STA\text{-}Q}^{(l)}$ denote the Local Spatiotemporal Adaptation input and output of the $l^{th}$ transformer block. Meanwhile,  self-attention $\mathrm{ST\text{-}MSA}$ operates on the spatial dimension that has merged spatiotemporal tokens along it to explore the local spatiotemporal relationships.
 
It's important to note that, unlike the query branch, the support branch is equipped with class text labels corresponding to support videos. This enables the Local Spatiotemporal Adaptation to transition into Local Multimodal Adaptation by incorporating text tokens in each frame specific to the FSAR support branch.  By introducing text priors, the module enhances its focus on action semantics during spatiotemporal modeling, thus achieving multimodal modeling.
Specifically, we input the labeled text description corresponding to each support set video $C^m \in \mathcal{C}$ into the CLIP text encoder to get text tokens $\textbf{F}_{\mathcal{T}}^m$ \big($\textbf{F}_{\mathcal{T}}^m \in \textbf{F}_{\mathcal{T}}$\big). The text encoder is frozen to avoid the extra computation cost and catastrophic forgetting phenomenon. To facilitate the fusion of multimodal data, we process the text features $\textbf{F}_{\mathcal{T}}^m \in \mathbb{R} ^ {1 \times D'}$ as follows:
 \begin{equation}\label{}
    \textbf{F}_{\mathcal{T}}^{MA} = \mathrm{Repeat} \left( \mathrm{FC}_{text} \left(\textbf{F}_{\mathcal{T}}^m \right) \right)
\end{equation}
where $\mathrm{FC}_{text} \in \mathbb{R} ^ {D' \times D}$ aims to align text tokens with video tokens in the feature dimension, and the $\mathrm{FC}_{text}$ weights are shared across all layers of the visual transformer. The $\mathrm{Repeat}$ operation duplicates text features $T$ times to obtain $\textbf{F}_{\mathcal{T}}^{MA} \in \mathbb{R} ^ {T \times 1 \times D}$. For the support set branch, given the global temporal token $ {\textbf{x}}_{TA}^{(l)} \in \mathbb{R}^{T \times 1 \times D}$, the input visual tokens $\textbf{z}^{(l-1)}\in \mathbb{R}^{T \times (N+1) \times D}$ and the text semantic token $\textbf{F}_{\mathcal{T}}^{MA} \in \mathbb{R}^{T \times 1 \times D}$, we concatenate these tokens together along the spatial dimension to obtain ${\textbf{z}}_{MA\text{-}S}^{(l-1)} = \left[\textbf{z}^{(l-1)}; \textbf{x}_{TA}^{(l)} ; \textbf{F}_{\mathcal{T}}^{MA}  \right] \in \mathbb{R}^{T \times (N+3) \times D} $, where $N$ denotes the total number of patches. However, the corresponding text labels for the videos are unknown for the query set branch, so we can only concatenate the input video features $\textbf{z}^{(l-1)}$ and temporal adapted features $\textbf{x}_{TA}^{(l)}$ to obtain ${\textbf{z}}_{STA\text{-}Q}^{(l-1)}=\left[ \textbf{z}^{(l-1)}; \textbf{x}_{TA}^{(l)} \right] \in \mathbb{R}^{T \times (N+2) \times D}$. Then, we feed ${\textbf{z}}_{MA\text{-}S}^{(l-1)}$ into Local Multimodal Adaptation to fuse spatiotemporal information with text semantic information as illustrated  in Fig.~\ref{fig:FgMA}(c), given by:
\begin{equation}\label{}
    {\textbf{z}}_{MA\text{-}S}^{(l)}={\textbf{z}}_{MA\text{-}S}^{(l-1)}+\mathrm{Adapter}\left( \mathrm{LM\text{-}MSA}\left(\mathrm{LN}\left({\textbf{z}}_{MA\text{-}S}^{(l-1)} \right) \right) \right)
\end{equation}
 where ${\textbf{z}}_{MA\text{-}S}^{(l-1)}$ and ${\textbf{z}}_{MA\text{-}S}^{(l)}$ denote the Local Multimodal Adaptation input and output of the $l^{th}$ ViT block. Self-attention $\mathrm{LM\text{-}MSA}$ operates on the spatial dimension, where multimodal tokens have been merged, to perform local spatiotemporal modeling and enhance attention to action-related semantics. 
 The  Local Spatiotemporal and Multimodal Adaptation share weight parameters, allowing query and support samples to be in the same feature space.

\subsubsection{\textbf{Joint Adaptation}}
\label{Joint_AD}
Lastly, we introduce Joint Adaptation, in which an Adapter is parallel to the MLP layer to tune the final representations jointly. Specifically, to ensure the consistency of each layer of the transformer block in the spatial dimension, we perform the $\mathrm{Select}$ operation on  $ {\textbf{z}}_{MA\text{-}S}^{(l)}$ and ${\textbf{z}}_{STA\text{-}Q}^{(l)}$, taking the first $N+1$ features in the spatial dimension of them. Joint adaptation can be computed as follows:
%  \begin{equation}\label{}
%  \mathbf{z}^{(l)}=\mathbf{z}_{MA\text{-}S}^{(l)}+\mathrm{MLP}\left( \mathrm{LN}\left( \mathbf{z}_{MA\text{-}S}^{(l)} \right) \right) +  s\cdot Adapter\left( \mathrm{LN}\left( \mathbf{z}_{MA\text{-}S}^{(l)} \right) \right)
% \end{equation}
\vspace{-3pt}
 \begin{equation}\label{}
\mathbf{z}^{(l)}=\begin{cases}
	\mathbf{z}_{MA\text{-}S}^{(l)}+\mathrm{MLP}\left( \mathrm{LN}\left( \mathbf{z}_{MA\text{-}S}^{(l)} \right) \right) + \\ \ \ \ \ \ \ \ \ \ \ \ \ r\cdot \mathrm{Adapter}\left( \mathrm{LN}\left( \mathbf{z}_{MA\text{-}S}^{(l)} \right) \right) \ \ \ \ if \ i=0 \,\,\\
	\mathbf{z}_{STA\text{-}Q}^{(l)}+\mathrm{MLP}\left( \mathrm{LN}\left( \mathbf{z}_{STA\text{-}Q}^{(l)} \right) \right) +\\ \ \ \ \ \ \ \ \ \ \ \ \ r\cdot \mathrm{Adapter}\left( \mathrm{LN}\left( \mathbf{z}_{STA\text{-}Q}^{(l)} \right) \right) \ \ \ if \ i=1 \,\, \\
\end{cases}
\end{equation}
where $i=0$ refers to the support set branch and $i=1$ refers to the query set branch. In this context, $r$ is a scaling factor that regulates the influence of the Adapter's output weight.

\begin{figure} [t!]
		\centering
	\includegraphics[width=0.9\linewidth,height=0.7\linewidth]{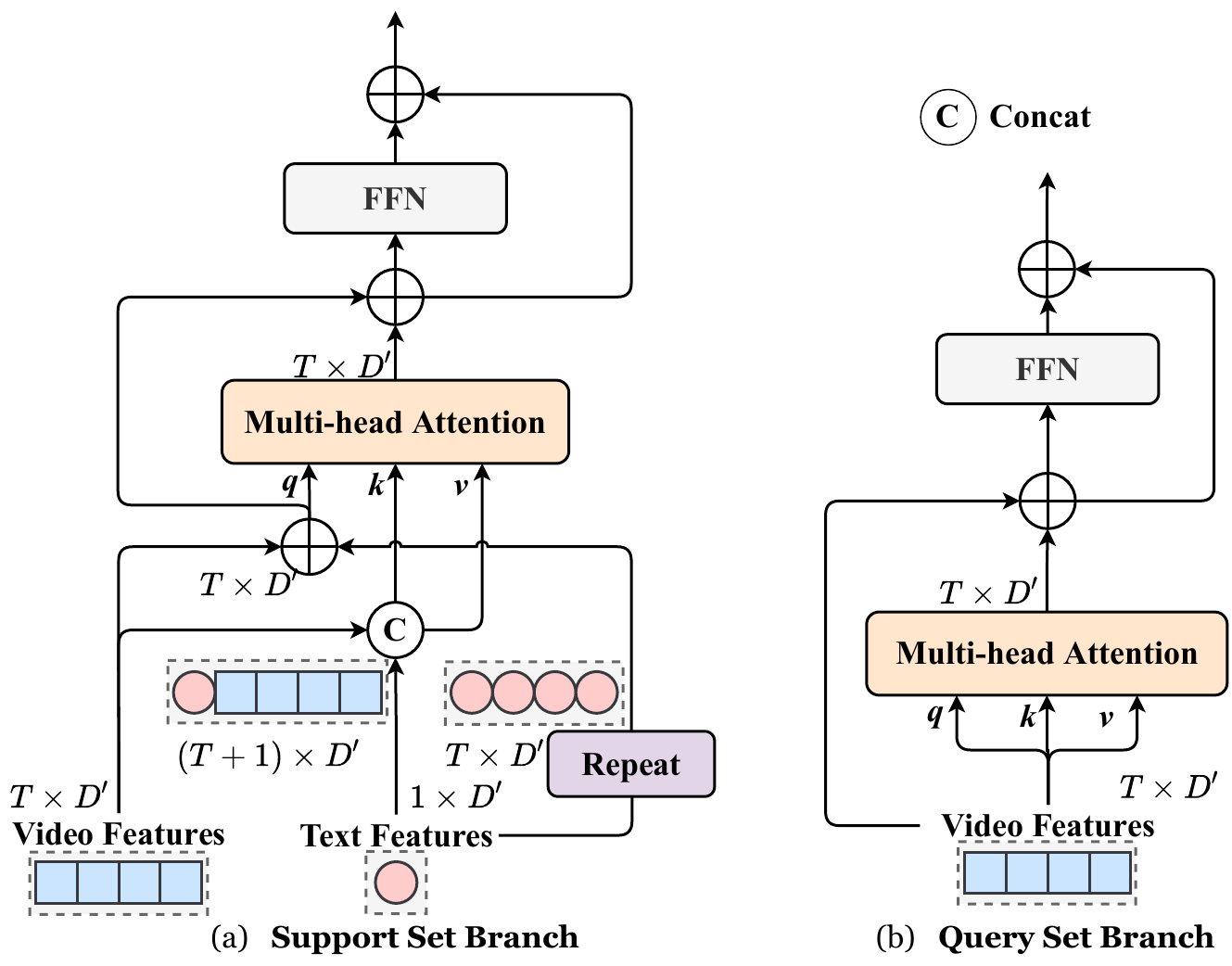}
 \vspace{-7pt}
		\caption{(a) and (b) respectively show the structure of the TPCM module for the support set and query set branch. $\oplus$ denotes element-wise summation.}
		\label{fig:TPCM}
  \vspace{-13pt}
	\end{figure}
 
\subsection{Text-guided Prototype Construction Module (TPCM)}
%  Over the few years, the success of multimodal methods in action recognition~\cite{wang2021actionclip, ni2022expanding, ju2022prompting, lin2022frozen, wasim2023vita} has demonstrated that it is possible to understand and represent
% the semantic information contained in the video more accurately by jointly modeling the video data and relevant textual information.
% To solve this problem and further completely leverage the powerful multimodal capabilities of CLIP,that can fully utilize semantic-level video-text features from CLIP to enhance the representation of video prototypes
In few-shot action recognition (FSAR), the quality of class prototype construction directly affects the performance of class prototype matching. Previous methods~\cite{cao2020few, zhu2018compound, zhang2020few, perrett2021temporal, thatipelli2022spatio, xing2023revisiting, wang2022hybrid, li2022hierarchical, huang2022compound} focus on using limited video features to construct class prototypes, which can easily lead to confusion among prototypes of similar categories.  Therefore, we design a Text-guided Prototype Construction Module (TPCM)  to enrich temporal and semantic representations at the prototype level by fully leveraging the powerful multimodal capabilities of CLIP. Specifically, for the support set branch, given the adapted features from FgMA $\textbf{F}_{\mathcal{S}}^m \in \textbf{F}_{\mathcal{S}}$ and the corresponding text features $\textbf{F}_{\mathcal{T}}^m \in \textbf{F}_{\mathcal{T}} $, we apply the cross-attention between them to utilize text features for guiding the construction of support class prototypes, resulting in $\widetilde{\textbf{F}_{\mathcal{S}}^m} \in \widetilde{\textbf{F}_{\mathcal{S}}}$. The process of obtaining the query-key-value triplets 
 $\textbf{q}_{\mathcal{S}}^m$, $\textbf{k}_{\mathcal{S}}^m$, $\textbf{v}_{\mathcal{S}}^m$  can be described as follows:
 \begin{equation}\label{}
 \textbf{q}_{\mathcal{S}}^m = \textbf{F}_{\mathcal{S}}^m + \mathrm{Repeat}\left( \textbf{F}_{\mathcal{T}}^m \right) \\
 \end{equation}
 \vspace{-10pt}
 \begin{equation}\label{}
  \textbf{k}_{\mathcal{S}}^m =  \textbf{v}_{\mathcal{S}}^m = \mathrm{Concat}\left( \left[ \textbf{F}_{\mathcal{S}}^m; \textbf{F}_{\mathcal{T}}^m \right]\right)
 \end{equation}
where $\textbf{F}_{\mathcal{S}}^m \in \mathbb{R} ^ {T \times D'}$, $\textbf{F}_{\mathcal{T}}^m \in \mathbb{R} ^ {1 \times D'}$, $\textbf{q}_{\mathcal{S}}^m \in \mathbb{R} ^ {T \times D'}$, $\textbf{k}_{\mathcal{S}}^m = \textbf{v}_{\mathcal{S}}^m  \in \mathbb{R} ^ {(T+1) \times D'}$, and $ \mathrm{Repeat}$ aims to copy $\textbf{F}_{\mathcal{T}}^m$ $T$ times. Then, we apply  the multi-head attention (MHA) and a feed-forward network (FFN) to obtain the enhanced support class prototype $\widetilde{\textbf{F}_{\mathcal{S}}^m} \in \mathbb{R} ^ {T \times D'}$ as shown in Fig.~\ref{fig:TPCM}(a), given by:
 \begin{equation}\label{}
{\bar{\textbf{F}}_{\mathcal{S}}^m} = \textbf{q}_{\mathcal{S}}^m + \mathrm{MHA}\left(\textbf{q}_{\mathcal{S}}^m, \textbf{k}_{\mathcal{S}}^m, \textbf{v}_{\mathcal{S}}^m \right)
 \end{equation}
 \begin{equation}\label{}
\widetilde{\textbf{F}_{\mathcal{S}}^m} = {\bar{\textbf{F}}_{\mathcal{S}}^m} + \mathrm{FFN}\left( {\bar{\textbf{F}}_{\mathcal{S}}^m} \right)
 \end{equation}
% where $\mathrm{MHA}$ consists of the layer normalization and a multi-head attention layer, and $\mathrm{FFN}$ consists of the layer normalization and an MLP layer.
Similarly, we perform the same operation on the query set videos to achieve the temporal enhancement at the prototype level, as shown in Fig.~\ref{fig:TPCM}(b). However, the difference is that $\textbf{q}_{\mathcal{Q}}^m=\textbf{k}_{\mathcal{Q}}^m=\textbf{v}_{\mathcal{Q}}^m=\textbf{F}_{\mathcal{Q}}^m \in \mathbb{R}^{T \times D'}$ since it does not have corresponding textual features. Note that the support and query set branches share the parameter weights of all modules to reduce computation costs while ensuring that query and support samples are in the same feature space. By incorporating the guidance of powerful multimodal information in constructing class prototypes, we can optimize intra-class and inter-class correlations of video features.

\subsection{Metric Loss and Predictions}
\label{prediction}
        Existing few-shot action recognition works~\cite{cao2020few, zhu2018compound, zhang2020few, perrett2021temporal, thatipelli2022spatio, xing2023revisiting, wang2022hybrid, li2022hierarchical, huang2022compound}, relying solely on visual information, classify a query video by comparing the temporally-aligned distances between the query video and the support set prototypes. The advent of the visual-language pre-training model CLIP enables query videos to be classified by matching not only with the prototypes of the support set (visual branch) but also with the corresponding text features of the support set (text branch), as shown in Fig.~\ref{fig:pipeline}. For the visual branch, given the enhanced class support prototype $\widetilde{\textbf{F}_\mathcal{S}^m} \in \widetilde{\textbf{F}_\mathcal{S}}$ and the query enhanced feature $\widetilde{\textbf{F}_q} \in \widetilde{\textbf{F}_{\mathcal{Q}}}$, the distance $D_{q,\mathcal{S}^m}$ can be calculated as:
 \begin{equation}\label{}
 D_{q,\mathcal{S}^m} =\mathcal{M}\left(\widetilde{\textbf{F}_q} ,\widetilde{\textbf{F}_\mathcal{S}^m}\right)  \\
 \end{equation}
where $\mathcal{M}$ denotes the temporal alignment metric, and $ D_{q,\mathcal{S}^m}\in D_{q,\mathcal{S}}$. Based on the distances $D_{q,\mathcal{S}} $, we can obtain the probability distribution over support classes $\textbf{p}_{\mathcal{Q}2\mathcal{S}}$ and use a standard cross-entropy loss $\mathcal{L}_{\mathcal{Q}2\mathcal{S}}$ to optimize the model parameters. For the text branch, given the adapted support set prototype feature ${\textbf{F}_\mathcal{S}^m} \in  {\textbf{F}_\mathcal{S}}$, adapted query feature ${\textbf{F}_q} \in {\textbf{F}_{\mathcal{Q}}}$, and corresponding text feature ${\textbf{F}_\mathcal{T}^m} \in  {\textbf{F}_\mathcal{T}}$, we apply global average pooling on temporal dimension to the features ${\textbf{F}_\mathcal{S}^m}$ and ${\textbf{F}_q}$ to obtain ${\textbf{F}_\mathcal{S}^{m\text{-}avg}}$ and  ${\textbf{F}_q^{avg}}$. To bring the pairwise representations of videos and labels closer to each other, we define symmetric similarities between the two modalities using cosine distances in the similarity calculation module, given by:
 \begin{equation}\label{}
s\left( {\textbf{F}_\mathcal{S}^{m\text{-}avg}},  {\textbf{F}_\mathcal{T}^m}\right) = \frac{\left< \mathbf{F}_{\mathcal{S}}^{m\text{-}avg},\mathbf{F}_{\mathcal{T}}^{m} \right>}{\left\| \mathbf{F}_{\mathcal{S}}^{m\text{-}avg} \right\| \left\| \mathbf{F}_{\mathcal{T}}^{m} \right\|} 
 \end{equation}
 \begin{equation}\label{}
 s\left( {\textbf{F}_q^{avg}},  {\textbf{F}_\mathcal{T}^m}\right) = \frac{\left< {\textbf{F}_q^{avg}},\mathbf{F}_{\mathcal{T}}^{m} \right>}{\left\|{\textbf{F}_q^{avg}} \right\| \left\| \mathbf{F}_{\mathcal{T}}^{m} \right\|}
 \end{equation}
where $s\left( {\textbf{F}_\mathcal{S}^{m\text{-}avg}},  {\textbf{F}_\mathcal{T}^m}\right)\in s\left( {\textbf{F}_\mathcal{S}^{avg}},  {\textbf{F}_\mathcal{T}}\right)$ and $s\left( {\textbf{F}_q^{avg}},  {\textbf{F}_\mathcal{T}^m}\right) \in s\left( {\textbf{F}_\mathcal{Q}^{avg}},  {\textbf{F}_\mathcal{T}}\right)$. Based on the cosine similarities $s\left( {\textbf{F}_\mathcal{S}^{avg}},  {\textbf{F}_\mathcal{T}}\right)$ and $s\left( {\textbf{F}_\mathcal{Q}^{avg}},  {\textbf{F}_\mathcal{T}}\right)$, we can obtain the softmax-normalized video-to-text similarity scores $\textbf{p}_{\mathcal{S}2\mathcal{T}}$ and  $\textbf{p}_{\mathcal{Q}2\mathcal{T}}$. Inspired by ActionCLIP~\cite{wang2023actionclip}, we define the Kullback–Leibler (KL) divergence as the video-text contrastive loss $\mathcal{L}_{\mathcal{S}2\mathcal{T}}$ and $\mathcal{L}_{\mathcal{Q}2\mathcal{T}}$. By optimizing contrastive loss, the CLIP model can be adapted to our FSAR task. Finally,  we integrate the losses of both the visual and textual branches, given by:
 \begin{equation}\label{}
 \mathcal{L} = \alpha\cdot   \frac{1}{2} \left( \mathcal{L}_{\mathcal{S}2\mathcal{T}} + \mathcal{L}_{\mathcal{Q}2\mathcal{T}}  \right)  + \left(1- \alpha\right) \cdot \mathcal{L}_{\mathcal{Q}2\mathcal{S}}
 \end{equation}
We also combine the query set video prediction distributions from both the visual and text branches, written as:
 \begin{equation}\label{}
 \textbf{p} = \alpha\cdot  \textbf{p}_{\mathcal{Q}2\mathcal{T}} + \left(1- \alpha\right)\cdot   \textbf{p}_{\mathcal{Q}2\mathcal{S}}
 \end{equation}
where $\alpha \in \left[0,1 \right]$ is an adjustable hyperparameter. The inference process MA-FSAR is summarized in Algorithm~\ref{inference}.
\begin{algorithm2e}[t]
\caption{The inference process of MA-FASR}
{\bf Input:} \\ $\mathcal{Q}$ (query set),  $\mathcal{S}$ (support set),  $\mathcal{C}$ (text descriptions) \vspace{8pt}

% ${\mathcal G} = ({\mathcal V},{\mathcal E}; {\mathcal T})$, where ${\mathcal T} = {\mathcal S}  \bigcup {\mathcal Q}$, ${\mathcal S} = \{({\bf x}_i, y_i)\}_{i=1}^{N \times K}$, ${\mathcal Q} =\{{\bf x}_i\}_{i=N \times K + 1}^{N \times K + T}$ \

{\bf Output:} \\ \textbf{p} (query set video prediction distribution)
% $\{{\hat y}_{i}\}_{i=N \times K + 1}^{N \times K + T}$\
\vspace{8pt}

{\bf Feature Extraction:} \\
 $\textbf{F}_{\mathcal{T}} = f_{text}\left(  \mathrm{Prompt}\left(\mathcal{C}\right)\right)$ \\
 $\textbf{F}_{\mathcal{S}} = \mathrm{FgMA}\left(\mathcal{S},\textbf{F}_{\mathcal{T}} \right)$, \ \ \ $\textbf{F}_{\mathcal{Q}} = \mathrm{FgMA}\left(\mathcal{Q}\right)$ 
   \vspace{8pt}

{\bf Prototype Construction:} \\
$\widetilde{\textbf{F}_\mathcal{S}} = \mathrm{TPCM}\left(\textbf{F}_{\mathcal{S}}, \textbf{F}_{\mathcal{T}}\right)$, \  \ \ $\widetilde{\textbf{F}_\mathcal{Q}} = \mathrm{TPCM}\left(\textbf{F}_{\mathcal{Q}}\right)$  \vspace{8pt}

{\bf Prototype Matching:} \\
$\textbf{F}_{\mathcal{Q}}^{avg}=\mathrm{AvgPool}\left(\textbf{F}_{\mathcal{Q}}, dim=1 \right)$ \\ $\textbf{p}_{\mathcal{Q}2\mathcal{T}} = \mathrm{Softmax}\left(\frac{\left< {\textbf{F}_{\mathcal{Q}}^{avg}},\mathbf{F}_{\mathcal{T}} \right>}{\left\|{\textbf{F}_{\mathcal{Q}}^{avg}} \right\| \left\| \mathbf{F}_{\mathcal{T}} \right\|}\right)$\\
 $\textbf{p}_{\mathcal{Q}2\mathcal{S}} = \mathrm{Softmax}\left(\mathcal{M}\left(\widetilde{\textbf{F}_{\mathcal{Q}}} ,\widetilde{\textbf{F}_\mathcal{S}}\right)  \right)$ \\
 $\textbf{p} = \alpha\cdot  \textbf{p}_{\mathcal{Q}2\mathcal{T}} + \left(1- \alpha\right)\cdot   \textbf{p}_{\mathcal{Q}2\mathcal{S}}$
 \label{inference}
\end{algorithm2e}

\section{Experiments}
\subsection{Experimental Setup}
\subsubsection{\textbf{Datasets}}
 Our method's performance is assessed on five datasets that can be classified into two categories: 1) spatial-related datasets, including Kinetics~\cite{carreira2017quo}, HMDB51~\cite{kuehne2011hmdb}, and UCF101~\cite{soomro2012ucf101}. 2) temporal-related datasets, including SSv2-Full~\cite{goyal2017something} and SSv2-Small~\cite{goyal2017something}. For the former, action recognition primarily relies on background information, with temporal information playing a minor role. In contrast, for the latter, the key to action recognition lies in effective temporal modeling. Referring to the previous setups~\cite{cao2020few, zhu2021closer, zhu2018compound} on Kinetics, SSv2-Full, SSv2-Small, we select 100 classes and divide them into 64/12/24 action classes as training/validation/testing classes. For UCF101 and HMDB51, we evaluate our method on the splits provided by~\cite{zhang2020few}.
\subsubsection{\textbf{Network Architectures}} \label{NA}
We choose CLIP~\cite{radford2021learning} as our backbone for efficient fine-tuning, where the visual encoder is ViT-B/32 or ViT-B/16, and the text encoder is a 12-layer, 512-wide transformer with eight attention heads. However, due to the previous works~\cite{cao2020few, zhu2018compound, zhang2020few, perrett2021temporal, thatipelli2022spatio, xing2023revisiting, wang2022hybrid, li2022hierarchical, huang2022compound}  using ResNet-50~\cite{he2016deep}  pre-trained on ImageNet~\cite{deng2009imagenet} as the backbone, we provide a version of utilizing pre-trained CLIP ResNet50 without the FgMA module as our visual encoder. Meanwhile, we set the bottleneck ratio of Adapters to 0.25 in the FgMA module (Sec.~\ref{FgMA_module}),  the same as AIM~\cite{yang2023aim}. And, we set the scaling factor $r$ to 0.5 on Joint Adaptation (Sec.~\ref{Joint_AD}). As for the adjustable hyperparameter $\alpha$ that controls metric loss and predictions (Sec.~\ref{prediction}), we set $\alpha$ to 0.75 for spatial-related datasets and 0.25 for temporal-related datasets. For the prompt templates of the text encoder, we discussed in Sec.~\ref{prompt_template}.  In training, a prompt template is randomly selected from 16 candidate templates for each video. The vector is obtained during inference by utilizing all 16 prompt templates as inputs and taking their average. For the temporal alignment metric $\mathcal{M}$, we choose OTAM~\cite{cao2020few} as our matching metric.
\begin{table*}[t!]
\centering
\caption{Comparison under 5-way k-shot settings on spatial-related benchmarks including HMDB51, UCF101, and Kinetics. The \textbf{boldfacen} and \underline{underline font} indicate the highest and the second highest results.  Note: * means our implementation. For Fine-tuning, ``Full'' indicates the full-parameter fine-tuning of the visual encoder, and ``PEFT'' indicates the parameter-efficient fine-tuning of the visual encoder. Methods with the CLIP pre-training all use textual input from the support set videos except AIM~\cite{yang2023aim}.}
\vspace{-7pt}
% \scriptsize
% \setlength{\tabcolsep}{0.5em}
\begin{tabular}{cccccccccc}
\hline
\rowcolor[HTML]{ECF4FF} & & & &
\multicolumn{2}{c}{\textbf{HMDB51}} & \multicolumn{2}{c}{\textbf{UCF101}}  & \multicolumn{2}{c}{\textbf{Kinetics}} \\
\rowcolor[HTML]{ECF4FF}\multirow{-2}{*}{\textbf{Method}}& \multirow{-2}{*}{\textbf{Reference}}&  \multirow{-2}{*}{\textbf{Pre-training}} & \multirow{-2}{*}{\textbf{Fine-tuning}} & \textbf{1-shot} & \textbf{5-shot} & \textbf{1-shot} & \textbf{5-shot} & \textbf{1-shot} & \textbf{5-shot}  \\ \hline
CMN++~\cite{zhu2018compound} & ECCV(18)& INet-RN50 &Full &- & - & - & -  & 57.3 & 76.0 \\
OTAM~\cite{cao2020few} & CVPR(20)& INet-RN50 &Full & 54.5 & 68.0 & 79.9 & 88.9  & 73.0 & 85.8 \\
TRX~\cite{perrett2021temporal} & CVPR(21)& INet-RN50 & Full&54.9* & 75.6 & 81.0* & 96.1  & 65.1* & 85.9 \\
STRM~\cite{thatipelli2022spatio}  & CVPR(22)& INet-RN50 &Full & 57.6* & 77.3 & 82.7* & 96.9  & 65.1* & 86.7 \\
HyRSM~\cite{wang2022hybrid} & CVPR(22)& INet-RN50 &Full&60.3& 76.0 & 83.9 & 94.7 & 73.7 & 86.1 \\
HCL~\cite{zheng2022few} & ECCV(22)& INet-RN50 &Full & 59.1 & 76.3 & 82.5 & 93.9  & 73.7 & 85.8 \\
Huang $et al.$~\cite{huang2022compound}& ECCV(22)& INet-RN50 &Full & 60.1 & 77.0 & 71.4 & 91.0  & 73.3 & 86.4 \\
Nguyen $et al.$~\cite{nguyen2022inductive}& ECCV(22)& INet-RN50 &Full & 59.6 & 76.9 & 84.9 & 95.9 & 74.3 & 87.4 \\
SloshNet~\cite{xing2023revisiting} & AAAI(23)&INet-RN50 &Full & 59.4 & 77.5 & 86.0 & 97.1  & 70.4 & 87.0 \\
MoLo (OTAM)~\cite{wang2023molo} & CVPR(23)& INet-RN50 &Full & 59.8 & 76.1 & 85.4 & 95.1  & 73.8 & 85.1 \\
GgHM ~\cite{xing2023boosting} & ICCV(23) & INet-RN50 &Full & 61.2 & 76.9 & 85.2 & 96.3 & 74.9 & 87.4  \\

\hline
CLIP-FSAR ~\cite{wang2023clip} & IJCV(23) & CLIP-RN50 & Full & 69.4 & 80.7 & 92.4 & 97.0 & 90.1 & 92.0 \\

CLIP-FSAR ~\cite{wang2023clip} & IJCV(23) & CLIP-ViT-B/16 & Full & 77.1 & 87.7 & \underline{97.0} & \underline{99.1} & \underline{94.8} & {95.4}
\\ 
AIM (OTAM)~\cite{yang2023aim} &ICLR(23) & CLIP-ViT-B/32 &  PEFT  &  74.7*&  83.1*& 92.3* & 96.8* &89.0* & 92.8* \\
AIM (OTAM)~~\cite{yang2023aim} &ICLR(23) & CLIP-ViT-B/16 &  PEFT  &  76.2* &  86.9* & 95.4*  & 98.2* &90.5* & 95.3*  \\ 
MVP-shot~\cite{qu2024mvp} & Arxiv(24) & CLIP-RN50 & Full  &  72.5&  82.5 & 92.2 & 97.6 & 90.0 & 93.2 \\
MVP-shot~\cite{qu2024mvp} & Arxiv(24) & CLIP-ViT-B/16 & Full  &  77.0&  88.1& 96.8 & 99.0 & 91.0 & 95.1\\
CLIP-CPM$^2$C~\cite{yang2023aim} &Arxiv(24) & CLIP-RN50 & Full  &  66.3&  81.2& 91.1 & 97.4 &88.6& 93.1 \\
CLIP-CPM$^2$C~\cite{yang2023aim} &Arxiv(24) & CLIP-ViT-B/16 & Full  & 75.9  & \textbf{88.0}  &   95.0 & 98.6 &  91.0 & \underline{95.5} \\
\hline
\textbf{MA-FSAR} & - & CLIP-RN50 & Full & 73.3 & 82.1 & 92.8 & 97.2 & 92.8 &93.0 \\
\textbf{MA-FSAR} & - & CLIP-ViT-B/32 & PEFT & \underline{77.3} & 83.9 & 95.0 & 98.7 & 93.5 &94.3 \\
\textbf{MA-FSAR} & - & CLIP-ViT-B/16 & PEFT & \textbf{83.4} & \underline{87.9} & \textbf{97.2} & \textbf{99.2} & \textbf{95.7} &\textbf{96.0} \\ \hline
\end{tabular}
\label{tab:spatial}
\vspace{-12pt}

% \vspace{-0.75em}
\end{table*}

\subsubsection{\textbf{Training and Inference}}
Following TSN~\cite{wang2016temporal}, we uniformly select 8 frames ($T$=8) of a video as the input
augmented with some fundamental techniques, such as random horizontal flipping, cropping, and color jitter in training, while only center crop in inference.  For training, SSv2-Full and SSv2-Small randomly sample 100,000 training episodes, and the other datasets randomly sample 10,000 training episodes. Meanwhile, we freeze the pre-trained CLIP and only fine-tune lightweight adapters during the training process when the visual encoder is ViT. If the visual encoder is ResNet-50, we only freeze the text encoder and fully fine-tune the visual encoder. Moreover, our framework uses the Adam~\cite{kingma2014adam} optimizer with the multi-step scheduler. As for inference, the average results of  10,000 tasks randomly sampled from the test sets in all datasets are reported in our experiments.
\vspace{-10pt}
\subsection{Results}
In this work, we chose OTAM~\cite{cao2020few} as our temporal alignment metric and maintained this default setting in all subsequent experiments. Methods with the CLIP pre-training except for AIM~\cite{yang2023aim} all use textual input from the support set videos.  Our approach reports results using three different visual encoders. The CLIP-RN50 model has a fully fine-tuned visual encoder since it does not have an Adapter structure. In contrast, the two ViT-B models only fine-tune lightweight adapters during the training process. Furthermore, AIM~\cite{yang2023aim} serves as the video classification framework, where we replace its classification head with our same matching head (OTAM) for a fair comparison.
\subsubsection{\textbf{Results on Spatial-related Datasets}}
For spatial-related datasets, action recognition primarily depends on background information, with temporal modeling playing a supplementary role. CLIP is the large foundation image pre-trained model that mainly relies on background information to recognize images. Therefore, fine-tuning CLIP on spatial-related datasets will result in a significant improvement in few-shot action recognition.  As shown in Tab.~\ref{tab:spatial}, even our CLIP-RN50 model significantly improves accuracy in any task setting compared to excellent methods (such as TRX~\cite{perrett2021temporal}, STRM~\cite{thatipelli2022spatio}, HyRSM~\cite{wang2022hybrid}, 
TADRNet~\cite{wang2023task},
SloshNet~\cite{xing2023revisiting}, MoLo~\cite{wang2023molo}, et al.) that use ImageNet pre-training. Compared to CLIP-FSAR~\cite{wang2023clip}, MVP-shot~\cite{qu2024mvp}, CLIP-CPM$^2$C~\cite{guo2024multi},  and AIM~\cite{yang2023aim}, which uses the same CLIP pre-training, our MA-FASR achieves better results across almost all datasets and task settings. Especially in 1-shot tasks, our method excels by integrating text features at the token level in FgMA. This integration guides the representation of video features with a more precise semantic focus.

\begin{table*}[t!]
\centering
\caption{Comparison under 5-way k-shot settings on temporal-related benchmarks including  SSv2-Small, and SSv2-Full. The \textbf{boldfacen} and \underline{underline font} indicate the highest and the second highest results.  Note: * means our implementation. For Fine-tuning, ``Full'' indicates the full-parameter fine-tuning of the visual encoder, and ``PEFT'' indicates the parameter-efficient fine-tuning of the visual encoder. Methods with the CLIP pre-training all use textual input from the support set videos except AIM~\cite{yang2023aim}.}
\vspace{-7pt}
% \scriptsize
% \setlength{\tabcolsep}{0.5em}
\begin{tabular}{cccccccc}
\hline
\rowcolor[HTML]{ECF4FF} & & & & 
\multicolumn{2}{c}{\textbf{SSv2-Small}} & \multicolumn{2}{c}{\textbf{SSv2-Full}}  \\
\rowcolor[HTML]{ECF4FF}\multirow{-2}{*}{\textbf{Method}} & \multirow{-2}{*}{\textbf{Reference}}&  \multirow{-2}{*}{\textbf{Pre-training}} & \multirow{-2}{*}{\textbf{Fine-tuning}} & \textbf{1-shot} & \textbf{5-shot} &\textbf{1-shot} & \textbf{5-shot} \\ \hline
CMN++~\cite{zhu2018compound} & ECCV(18)& INet-RN50 &Full &34.4 & 43.8 & 36.2 &48.8    \\
OTAM~\cite{cao2020few} & CVPR(20)& INet-RN50 &Full &36.4 &48.0  & 42.8  & 52.3  \\
TRX~\cite{perrett2021temporal} & CVPR(21)& INet-RN50 & Full & 36.0* & 56.7* & 42.0* & 64.6  \\
STRM~\cite{thatipelli2022spatio}  & CVPR(22)& INet-RN50 &Full & 37.1* & 55.3* & 43.1* & 68.1 \\
HyRSM~\cite{wang2022hybrid} & CVPR(22)& INet-RN50 &Full & 40.6 & 56.1 & 54.3 & 69.0 \\
HCL~\cite{zheng2022few} & ECCV(22)& INet-RN50 &Full & 38.7  & 55.4   & 47.3 & 64.9  \\
Huang $et al.$~\cite{huang2022compound}& ECCV(22)& INet-RN50 &Full &38.9  &61.6  & 49.3 & 66.7  \\
Nguyen $et al.$~\cite{nguyen2022inductive}& ECCV(22)& INet-RN50 &Full &-  &-  & 43.8 & 61.1 \\
SloshNet~\cite{xing2023revisiting} & AAAI(23)&INet-RN50 &Full & -  & -  & 46.5 & 68.3   \\
MoLo (OTAM)~\cite{wang2023molo} & CVPR(23)& INet-RN50 &Full & 41.9 & 56.2 & 55.0 & 69.6  \\
GgHM ~\cite{xing2023boosting} & ICCV(23) & INet-RN50 &Full & - & - & 54.5 & 69.2  \\
\hline
CLIP-FSAR ~\cite{wang2023clip} & IJCV(23) & CLIP-RN50 & Full & 52.1 & 55.8 & 58.7 & 62.8  \\
CLIP-FSAR ~\cite{wang2023clip} & IJCV(23) & CLIP-ViT-B/16 & Full & 54.6 & 61.8 & 62.1 & {72.1}\\
AIM (OTAM)~\cite{yang2023aim} &ICLR(23) & CLIP-ViT-B/32 &  PEFT  & 52.0* & 61.6* & 61.4* & 63.9*  \\
AIM (OTAM)~\cite{yang2023aim} &ICLR(23) & CLIP-ViT-B/16 &  PEFT  & {53.9}* & \underline{63.5}* & \underline{62.7}*  & {72.1}* 
\\ 
MVP-shot~\cite{qu2024mvp} & Arxiv(24) & CLIP-RN50 & Full & 
51.2 & 57.0 & -& - \\
MVP-shot~\cite{qu2024mvp} & Arxiv(24) & CLIP-ViT-B/16 & Full & 
55.4 & 62.0 & -& - \\
CLIP-CPM$^2$C~\cite{qu2024mvp} & Arxiv(24) & CLIP-RN50 & Full & 
51.5 & 57.1 & 58.0& 64.0 \\
CLIP-CPM$^2$C~\cite{qu2024mvp} & Arxiv(24) & CLIP-ViT-B/16 & Full & 
52.3 & 62.6 & 62.1& \textbf{72.8} \\

\hline
\textbf{MA-FSAR} & - & CLIP-RN50 & Full & 52.5& 57.8  &58.8 & 63.6\\
\textbf{MA-FASR} & - & CLIP-ViT-B/32 & PEFT & \underline{56.5} & 62.3  & 61.9 & 64.5 \\
\textbf{MA-FSAR} & - & CLIP-ViT-B/16 & PEFT & \textbf{59.1} & \textbf{64.5} &\textbf{63.3} & \underline{72.3} \\ \hline
\end{tabular}
\label{tab:temporal}
\vspace{-12pt}
\end{table*}

\subsubsection{\textbf{Results on Temporal-related Datasets}}
For temporal-related datasets, the key to action recognition is temporal relationship understanding. The performance improvement from CLIP's strong pre-training is less significant than those for spatial-related datasets. However, our model continues to demonstrate excellent results, attributed to our comprehensive global and local temporal modeling within FgMA at the token level and our temporal relationship enhancement at the prototype level.  We report three model results using different visual encoders as shown in Tab.~\ref{tab:temporal}. Compared to the baseline OTAM~\cite{cao2020few}, our MA-FSAR using CLIP-RN50 as the visual encoder can bring 16.1\%, 16.0\% performance improvements in the 1-shot task, and 9.8\%, 11.3\% accuracy gains in the 5-shot task of SSv2-Small and SSv2-Full, respectively.  Meanwhile, our CLIP-RN50 model outperforms all other methods using ResNet-50 as the visual encoder in 1-shot tasks across all temporal-related datasets, highlighting the effectiveness of our TPCM module design.  Compared to CLIP-FSAR~\cite{wang2023clip}, MVP-shot~\cite{qu2024mvp}, CLIP-CPM$^2$C~\cite{guo2024multi},  and AIM~\cite{yang2023aim}, which uses the same CLIP pre-training and temporal alignment metric, our method has a significant performance improvement, especially in 1-shot tasks.  For the SSv2-Small dataset, even our ViT-B/32 model can perform better than any CLIP-based ViT-B/16 model, demonstrating the superiority of our FgMA design.
\vspace{-10pt}
\subsection{Ablation Study}
we employ CLIP-ViT-B/32 as our visual encoder, as the default setting for subsequent ablation studies. Ablation experiments ~\ref{Components_1}, ~\ref{Components_2}, ~\ref{CPMandSp}, ~\ref{cpPCM}, ~\ref{CPMetric}, ~\ref{SM}, and ~\ref{FullandAd} demonstrate the effectiveness of our module design; ~\ref{prompt_template} explores the impact of different prompt templates on our MA-FSAR; ~\ref{zero-shot perf} presents the zero-shot performance; ~\ref{transfer_nums} and ~\ref{Costs} highlight our MA-FSAR's training efficiency;  ~\ref{Hyperparameters} discuss the hyperparameters in our model; ~\ref{Visualization} provides visualizations of attention maps.

\subsubsection{\textbf{Impact of The Proposed Components}} \label{Components_1}
To assess the impact of each module (\textit{i.e.}, FgMA and TPCM) in our method, we conduct experiments under 5-way 1-shot settings on SSv2-Small and SSv2-Full. As indicated in Tab.~\ref{components}, we confirm the effectiveness of each component. The multimodal baseline freezes all the learnable weights without extra modules.  Specifically, in comparison to the baseline, the FgMA module results in accuracy improvements of 13.5\% and 16.3\% on SSv2-Small and SSv2-Full, respectively, while the TPCM module yields gains of 16.9\% and 19.6\% on the two datasets. The combination of all modules produces the most favorable outcomes, leading to accuracy gains of 27.7\% and 31.7\% on SSv2-Small and SSv2-Full, respectively, over the baseline.
\begin{table}[t!]
\centering
\caption{The impact of proposed modules on SSv2-Small and SSv2-Full in the 5-way 1-shot task.} 
\vspace{-7pt}
{\begin{tabular}{cccc}
\hline
\rowcolor[HTML]{FFCCC9}
    \textbf{FgMA} & \textbf{TPCM}   & \textbf{SSv2-Small} & \textbf{SSv2-Full} \\ \hline
    $\usym{2717}$ 	 &  $\usym{2717}$     & 28.8 &  30.2 \\ 
    $\usym{2717}$ 	 &  $\usym{2713}$      & 42.3 &  46.5\\  
    $\usym{2713}$ 	 &  $\usym{2717}$    & 45.7 &  49.8 \\  
    $\usym{2713}$ &  $\usym{2713}$       & 56.5 & 61.9\\ \hline
\end{tabular}}
\label{components}
\vspace{-10pt}
\end{table}

\subsubsection{\textbf{Effectiveness of The Adaptation Components}} \label{Components_2}
To demonstrate the effectiveness of our proposed multi-type adaptation in FgMA, we compare our method to two baselines. The first baseline represents a frozen space-only model without any adaptation, wherein all the trainable parameters of CLIP encoders are frozen, but the TPCM module is not included. The second baseline involves fully fine-tuning the visual encoder without any adaptation. As presented in Tab.~\ref{Adapter}, the fine-tuned visual-only model demonstrates a 12.4\% performance improvement over the first baseline but with an increase in tunable parameters from 3.15M to 90.99M. Our approach seeks to introduce lightweight adapters into a fully frozen visual model, maintaining the integrity of the pre-trained weights, to achieve superior performance compared to a fully fine-tuned model. In Tab.~\ref{Adapter}, after the Local Multimodal/Spatiotemporal Adaptation, the frozen model achieves a comparable performance with the full fine-tuned visual-only model (54.0 \textit{vs.} 53.9), while utilizing less than one-tenth of the parameter count of the latter (7.94M \textit{vs.} 90.99M). Note that since there is no injection of the global temporal token, this adaptation cannot perform temporal modeling.  Upon adding 
the Global Temporal and Joint Adaptation, they yield 1.9\% and 0.6\% performance improvements. Our final model delivers a 2.6\% accuracy improvement compared to the fine-tuned visual-only model, yet with only one-fifth of the tunable parameters.

\begin{table}[t!]
\centering
\caption{Effectiveness of the Adapter components in the SSv2-Small 5-way 1-shot task. LMA/LSTA, GTA, and JA indicate the Local Multimodal/Spatiotemporal Adaptation, Global Temporal Adaptation, and Joint Adaptation.} 
\vspace{-7pt}
\scalebox{1.0}{\begin{tabular}{cccc}\hline
\rowcolor[HTML]{FFCCC9} 
& & {\textbf{Tunable}}&\\ 
\rowcolor[HTML]{FFCCC9}\multirow{-2}{*}{$\textbf{Method}$} & \multirow{-2}{*}{$\textbf{Params}$} & \textbf{Params} &\multirow{-2}{*}{$\textbf{Acc}$}  \\ \hline
% \hline
% \rowcolor[HTML]{FFCCC9} & &
% \multicolumn{2}{c|}{Method} & \multicolumn{2}{c|}{Param(M)} & \multicolumn{2}{c|}{Tunable}    &  \multicolumn{2}{c|}{Acc}  \\ \hline
   Frozen	 & 154.4M  & 3.2M  & 42.3  \\
  Fine-tuned visual-only  &  154.4M  & 91.0M & 53.9 \\ \hline
    Frozen  + LMA/LSTA & 159.2M & 7.9M & 54.0 \\ 
    \ \ \ \ \ + GTA & 166.3M &15.1M & 55.9\\ 
    \ \ \ \ \  + JA &169.8M & 18.5M & 56.5\\ \hline
\end{tabular}}
\label{Adapter}
\vspace{-10pt}
\end{table}

\begin{table}[t!]
\centering
\caption{Effectiveness Comparison Between Local Multimodal and Spatiotemporal Adaptation (LMA/LSTA) in 5-way 1-shot task.} 
\vspace{-7pt}
{\begin{tabular}{ccc}
\hline
\rowcolor[HTML]{FFCCC9}
    \textbf{Method} & \textbf{Dataset} &   \textbf{Acc}  \\ \hline
Double  LSTA & SSv2-Small	      & 55.9  \\
    \textbf{LSTA and LMA} 	 & SSv2-Small    & \textbf{56.5} \\ \hline
 Double   LSTA & SSv2-Full     & 61.2\\
    \textbf{LSTA and LMA} & SSv2-Full     & \textbf{61.9}\\ \hline
\end{tabular}}
\label{MAvsSA}
\vspace{-10pt}
\end{table}

\begin{table}[t!]
\centering
\caption{Comparisons of Different Prototype Construction Methods. } 
\vspace{-7pt}
{\begin{tabular}{ccc}
\hline
\rowcolor[HTML]{FFCCC9}
    \textbf{Method} & \textbf{Visual Encoder}   & \textbf{Acc}  \\  
    Unimodal Transformer 	 &  CLIP-ViT-B/32     & 47.9  \\  
    Multimodal Transformer 	 &  CLIP-ViT-B/32    & 54.7 \\  
    \textbf{TPCM} &  CLIP-ViT-B/32       & \textbf{56.5} \\ \hline
\end{tabular}}
\label{TPCM}
\vspace{-15pt}
\end{table}

\begin{table*}[ht!]
\centering
\caption{Method effectiveness on different temporal alignment metrics on SSv2-Small and Kinetics in the 5-way 1-shot task. And effectiveness Comparisons between the Unimodal model  and the Multimodal model.} 
\vspace{-7pt}
{\begin{tabular}{ccccc}
\hline
\rowcolor[HTML]{FFCCC9}
    \textbf{Temporal Alignment Metric} &  \textbf{Model Modality} & \textbf{Pre-training} & \textbf{Kinetics} & \textbf{SSv2-Small}  \\ \hline
      OTAM~\cite{cao2020few}&  Unimodal  & INet-ViT-B/32 &  75.8 &  38.2  \\ 
      OTAM~\cite{cao2020few}&  Unimodal  & CLIP-ViT-B/32 & 83.7  & 44.8 \\ 
      \textbf{MA-FSAR (OTAM)}&   Multimodal  & CLIP-ViT-B/32  &  \textbf{93.5} & \textbf{56.5}           \\ \hline
       TRX~\cite{perrett2021temporal}&  Unimodal  & INet-ViT-B/32 &  67.2 &  37.3  \\ 
      TRX~\cite{perrett2021temporal}&  Unimodal  & CLIP-ViT-B/32 &  82.8  & 42.7 \\ 
      \textbf{MA-FSAR (TRX)}&   Multimodal  & CLIP-ViT-B/32  &  \textbf{92.8} & \textbf{52.4}           \\ \hline
         Bi-MHM~\cite{wang2022hybrid}&  Unimodal  & INet-ViT-B/32 &  75.2 &  39.5  \\ 
      Bi-MHM~\cite{wang2022hybrid}&  Unimodal  & CLIP-ViT-B/32 & 83.2   &  45.5 \\ 
     \textbf{MA-FSAR (Bi-MHM)}&   Multimodal  & CLIP-ViT-B/32  &  \textbf{93.2} & \textbf{56.9}           \\ \hline
       
\end{tabular}}
\label{Metrics}
\vspace{-10pt}
\end{table*}

\begin{table*}[h!]
\centering
\caption{Effectiveness comparison between full fine-tuning and adaptation on SSv2-Small and SSv2-Full in the 5-way 1-shot task. "Memory (G)" refers to the amount of video memory usage, and "Time (Hours)" indicates the time required to train 10,000 tasks, measured in hours on a single RTX3090.} 
\vspace{-7pt}

{\begin{tabular}{cccccc}
\hline
\rowcolor[HTML]{FFCCC9}
    \textbf{Method} & \textbf{Dataset} & \textbf{Tunable Param (M)}& \textbf{Memory (G)} & \textbf{Time (Hours)}  & \textbf{Acc}  \\ \hline
    Full fine-tuning & SSv2-Small	 & 90.99    & 13.5 & 3.4 & 53.9  \\
    \textbf{Adaptation} 	 & SSv2-Small &  \textbf{18.54} & \textbf{11.9} &  \textbf{3.0}  & \textbf{56.5} \\ \hline
    Full fine-tuning & SSv2-Full &    90.99  & 13.5  & 3.4  & 59.1\\
    \textbf{Adaptation} & SSv2-Full &  \textbf{18.54}  & \textbf{11.9}  &  \textbf{3.0}    & \textbf{61.9}\\ \hline
\end{tabular}}
\label{PEFT}
\vspace{-15pt}
\end{table*}

\begin{table}[t!]
\caption{Comparison experiments of different prompt templates on SSv2-Small, Kinetics in the 5-way 1-shot task. ``Mixture'' refers to using 16 different prompt templates, then summing and averaging the outputs of 16 text tokens during inference.} 
\vspace{-7pt}
\centering
\scalebox{0.95}{\begin{tabular}{cccc}
\hline
\rowcolor[HTML]{FFCCC9}
    \textbf{Prompt template} & \textbf{SSv2-Small}& \textbf{Kinetics}   \\ \hline
   {\small``\texttt{[CLS]}''} & 56.1 & 92.0\\\hline
    {\small``\texttt{a photo of action [CLS]}''} & 55.3 &92.7 \\\hline
        \textbf{Mixture}  & \textbf{56.5} & \textbf{93.5}\\\hline

\end{tabular}}
\label{prompt}
\vspace{-10pt}
\end{table}

\begin{table}[t!]
\caption{The list of 16 different prompt templates.} 
\vspace{-7pt}

\centering
\small{
\scalebox{1.0}{\begin{tabular}{c}
\hline
\rowcolor[HTML]{FFCCC9}
    \textbf{Prompt template}   \\ \hline
    {``\texttt{[CLS]}''}\\
    {``\texttt{a photo of action [CLS]}''} \\
       ``\texttt{a picture of action [CLS]}'' \\
        ``\texttt{Human action of [CLS]}'' \\
   ``\texttt{[CLS], an action}'' \\
   ``\texttt{[CLS] this is an action}'' \\
    ``\texttt{[CLS], a video of action}'' \\
   ``\texttt{Playing action of [CLS]}'' \\
   ``\texttt{Playing a kind of action, [CLS]}'' \\
   ``\texttt{Doing a kind of action, [CLS]}'' \\
   ``\texttt{Look, the human is [CLS]}'' \\
   ``\texttt{Can you recognize the action of [CLS]}'' \\
   ``\texttt{Video classification of [CLS]}'' \\
   ``\texttt{A video of [CLS]}'' \\
    ``\texttt{The man is [CLS]}'' \\
   ``\texttt{The woman is [CLS]}'' \\ \hline

\end{tabular}}}
\normalsize
\label{prompt_list}
\vspace{-10pt}
\end{table}

\begin{table}[t!]
\caption{Comparison of 5-way zero-shot performance with CLIP on spatial-related datasets (Kinetics, HMDB51, and UCF101) and temporal-related dataset (SSv2-Small). } 
\vspace{-7pt}
\centering
\scalebox{0.85}{\begin{tabular}{ccccccc}
\hline
\rowcolor[HTML]{FFCCC9}
    \textbf{Method}  & \textbf{SSv2-Small}& \textbf{Kinetics}  & \textbf{HMDB51} & \textbf{UCF101}  \\ \hline
    CLIP-Frezze  & 28.8 & 92.8 & 63.3 & 87.1\\
    CLIP-FSAR~\cite{wang2023clip}   & 44.5   & 92.9 & 72.1  & 90.1 \\
    AIM~\cite{yang2023aim} & 46.7 &  \textbf{93.0} &72.6 & 88.7 \\ \hline
       \textbf{MA-FSAR}  & \textbf{47.1} & \textbf{93.0} & \textbf{73.2} & \textbf{89.9}\\\hline
\end{tabular}}
\label{zero}
\vspace{-15pt}
\end{table}

\begin{table}[t!]
\centering
\caption{Comparison of different methods for the number of training tasks on SSv2-Small and Kinetics in the 5-way 1-shot task. MA-FSAR's temporal alignment metric is OTAM~\cite{cao2020few}. Note: ``Num'' indicates the number of training iterations.} 
\vspace{-7pt}
\scalebox{0.9}{\begin{tabular}{ccccc}
\hline
\rowcolor[HTML]{FFCCC9}
    \textbf{Method} & \textbf{Dataset} & \textbf{Pre-training}& \textbf{Num}  & \textbf{Acc}  \\ \hline
    OTAM~\cite{cao2020few} & SSv2-Small	 & INet-ViT-B/32    & 80000 & 38.2  \\
    HYRSM~\cite{wang2022hybrid} & SSv2-Small & INet-ViT-B/32 & 75000 & 40.4 \\
    TRX~\cite{perrett2021temporal} & SSv2-Small & INet-ViT-B/32 & 80000 & 37.3 \\
    \textbf{MA-FSAR}& SSv2-Small & CLIP-ViT-B/32 & \textbf{20000} & \textbf{56.5}\\ \hline
    OTAM~\cite{cao2020few} & Kinetics	 & INet-ViT-B/32    & 10000 & 83.7  \\
    HYRSM~\cite{wang2022hybrid} &  Kinetics & INet-ViT-B/32 & 10000 & 83.2 \\
    TRX~\cite{perrett2021temporal} &  Kinetics & INet-ViT-B/32 & 10000 & 82.8 \\
    \textbf{MA-FSAR}& Kinetics & CLIP-ViT-B/32 & \textbf{1000} & \textbf{93.5}\\ \hline
   
\end{tabular}}
\label{Num}
\vspace{-10pt}
\end{table}

\subsubsection{\textbf{Comparison Between Local Multimodal and Spatiotemporal Adaptation}} \label{CPMandSp}
To ensure a fair comparison of Local Multimodal and Spatiotemporal Adaptation (LMA/LSTA), we conduct experiments on the 5-way 1-shot task of SSv2-Small and SSv2-Full. As outlined in Sec.~\ref{MA}, the distinction between the two adaptations lies in the inclusion or exclusion of text features for self-attention with spatiotemporal features in support videos. The results presented in Table ~\ref{MAvsSA} demonstrate that using LMA instead of LSTA leads to performance improvements of 0.6\% and 0.7\% on SSv2-Small and SSv2-Full, respectively.  These findings illustrate the efficacy of enhancing the semantic representation of visual tokens by integrating textual tokens within the Adapter (Sec.~\ref{FgMA_module}).

\subsubsection{\textbf{Comparisons of Different Prototype Construction Methods}} \label{cpPCM}
To demonstrate the effectiveness of our proposed module and to compare the performance of various methods for prototype construction, we conduct experiments in the SSv2-Small 5-way 1-shot task. The first baseline unimodal transformer indicates the features $\textbf{F}_\mathcal{S}$ and $\textbf{F}_\mathcal{Q}$ performing self-attention on the temporal dimension. Contrasting this, the second baseline (CLIP-FSAR\cite{wang2023clip}) differs in that the text features  $\textbf{F}_\mathcal{T}$ are stacked along the temporal dimension before self-attention is conducted on the support features $\textbf{F}_\mathcal{S}$. We set all the layers of the transformer to be one. As shown in Tab.~\ref{TPCM}, our TPCM module brings 8.6\% and 1.8\% performance improvements compared to the unimodal transformer and multimodal transformer on SSv2-Small, respectively. These experimental results underscore the TPCM module's considerable efficacy in leveraging textual information as guidance to integrate visual and textual features at the prototype level effectively, resulting in the creation of more robust class prototype representations.

\begin{table}[t!]
\centering
\caption{Effect of the scaling factor $r$.} 
\vspace{-7pt}

% \small 

{\begin{tabular}{cccccc}
\hline
\rowcolor[HTML]{FFCCC9}
    \textbf{scaling factor $r$} & \textbf{0}   & \textbf{0.25} &\textbf{0.5} &\textbf{0.75} &\textbf{1} \\ \hline
   Kinetics   &   93.1  &  93.3  & \textbf{93.5} &  93.4 & 93.4  \\ 
    SSv2-Small	& 55.8 &  56.2 &   \textbf{56.5} & 56.3 & 56.2  \\ \hline 
    \end{tabular}}
\label{scale_factor_r}
\vspace{-10pt}
\normalsize
\end{table}

\begin{table}[t!]
\centering
\caption{Effect of the loss factor $\alpha$.} 
\vspace{-7pt}

% \small 

{\begin{tabular}{cccccc}
\hline
\rowcolor[HTML]{FFCCC9}
     \textbf{loss factor $\alpha$} & \textbf{0}   & \textbf{0.25} &\textbf{0.5} &\textbf{0.75} &\textbf{1} \\ \hline 
        Kinetics   &  91.7   &  92.5  & 92.9 &  \textbf{93.5} & 93.0  \\ 
    SSv2-Small	& 56.2 &  \textbf{56.5} &  54.9  & 52.7 & 47.1  \\ \hline
    \end{tabular}}
\label{loss_factor_a}
\vspace{-10pt}
\normalsize
\end{table}

\subsubsection{\textbf{MA-FSAR Effectiveness on Different Temporal Alignment Metrics}} \label{CPMetric}
We conduct the experiments using different temporal alignment metrics on the 5-way 1-shot task of Kinetics and SSv2-Small to demonstrate that our model is plug-and-play, capable of being integrated with any common matching metric.  We adopt three different temporal alignment metrics, including OTAM~\cite{cao2020few},  TRX~\cite{perrett2021temporal}, and Bi-MHM~\cite{wang2022hybrid}. As displayed in Tab.~\ref{Metrics}, our method showcases its ability to adapt to any temporal alignment metric, with the final accuracies closely correlated to the performance of the chosen metric. Moreover, regardless of the temporal alignment metric employed, our MA-FSAR consistently achieves the most outstanding performance compared to the baselines, providing compelling evidence for the superiority of our model.
%  and achieve significantly better results than their respective baselines, which shows the superiority of our model.

\subsubsection{\textbf{Unimodal Model vs. Multimodal Model}} \label{SM}
To compare the performance of the unimodal and multimodal models, we experiment with various pre-training and model modalities in the 5-way 1-shot task on Kinetics and SSv2-Small, evaluating them on multiple matching metrics. We provide two baselines for each metric: an ImageNet~\cite{deng2009imagenet} pre-trained unimodal model and a CLIP pre-trained unimodal model, with all baseline models' visual encoders fully fine-tuned. As depicted in Tab.~\ref{Metrics}, using a CLIP pre-trained unimodal model shows some performance improvements compared to the ImageNet pre-trained model, albeit relatively limited. However, when using our proposed MA-FSAR multimodal model, a significant performance improvement is observed in both datasets. Specifically, our MA-FSAR consistently achieves a minimum accuracy improvement of 15\% over the ImageNet pre-trained unimodal model and  10\% over the CLIP pre-trained unimodal model on two datasets. These results not only highlight the significance of text features for few-shot action recognition but also prove the effectiveness of our method.

% However, the parameter count of our MA-FSAR is only one-fifth of the full fine-tuned model, which demonstrates the efficiency of our model.
\begin{figure*} [h!]
	\centering
	\includegraphics[width=\linewidth, height=0.8\linewidth]{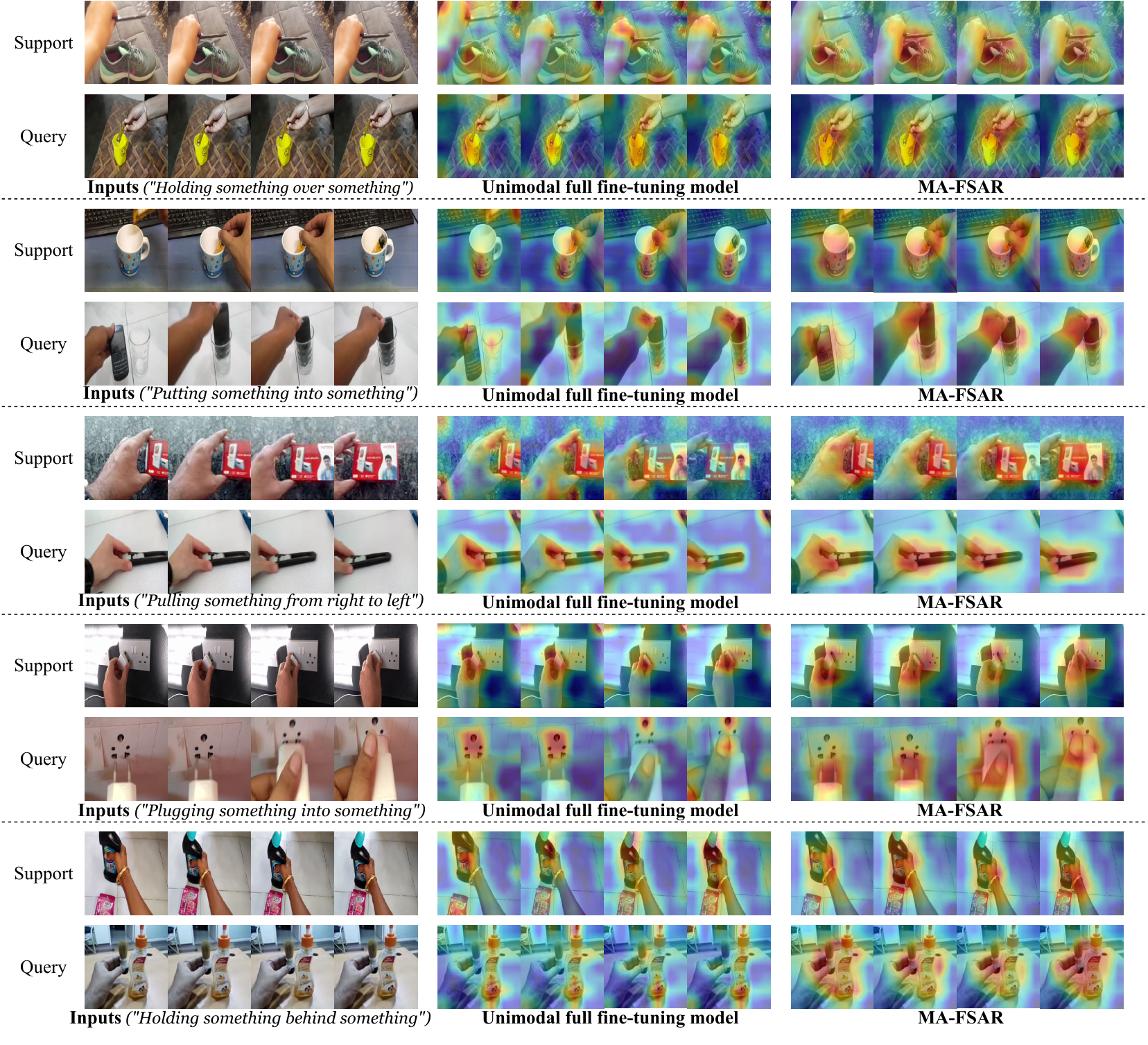}	\vspace{-14pt}
	\caption{Attention visualization of our MA-FSAR in the SSv2-Small 5-way 1-shot task. Corresponding to the original RGB images(left), the attention maps of the unimodal full fine-tuning model (middle) are compared to the attention maps with our MA-FSAR (right). The temporal alignment metric is OTAM~\cite{cao2020few}.}
	\label{fig:vis}
	\vspace{-10pt}
\end{figure*}

\subsubsection{\textbf{Full Fine-tuning vs. Adaptation}} \label{FullandAd}
We conduct experiments in the SSv2-Small 5-way 1-shot task to make a fair comparison between full fine-tuning and adaptation, which demonstrates the effectiveness of the FgMA module we propose. As shown in Tab.~\ref{PEFT}, our adaptation method leads to accuracy improvements of 2.6\% and 2.8\% on SSv2-Small and SSv2-Full, respectively, over the full fine-tuning model. Our adaptation method implements multimodal fusion and temporal modeling, while the full fine-tuning method does not achieve this. However, our method has only one-fifth (18.54M \textit{vs.} 90.99M) of tunable parameters compared to the full fine-tuning method, which requires 1.6G (11.9G \textit{vs.} 13.5G) less memory usage, and takes 0.4 hours (3.0H vs. 3.4H) less time to train for 10,000 tasks on a single RTX3090. The experimental results highlight that our MA-FSAR is fast, efficient, and has low training costs.

\subsubsection{\textbf{Comparison of different prompt templates}}
\label{prompt_template}
To explore the impact of different prompt templates on recognition performance, we conduct 5-way 1-shot experiments on Kinetics and SSv2-Small. Here, {\small``\texttt{[CLS]}''} refers to no prompt template, and {\small``\texttt{a photo of action [CLS]}''} is a commonly used prompt template. As illustrated in Tab.~\ref{prompt}, different datasets exhibit distinct preferences for prompt templates. For example, SSv2-Small performs better using {\small``\texttt{[CLS]}''}, while Kinetics shows the opposite trend. To address this, we introduce a mixed prompt template, which involves utilizing  16 different prompt templates and then summing and averaging the outputs of 16 text tokens during inference. The list of these 16 different prompt templates is shown in Tab.~\ref{prompt_list}. In Tab.~\ref{prompt}, our mixture method balances the preferences of various datasets for prompt templates, achieving optimal performance on each dataset.

\begin{table}[t!]
\centering
\caption{Comparsion of training costs between the baseline methods OTAM~\cite{cao2020few}, CLIP-FSAR~\cite{wang2023clip}, and our MA-FSAR.}
\vspace{-8pt}
% \small 
\scalebox{0.85}{\begin{tabular}{cccccc}
\hline
\rowcolor[HTML]{FFCCC9}

    \textbf{Method} &\textbf{Encoder} & \textbf{Params (tune)}   & \textbf{Memory} &\textbf{Time} &\textbf{Acc} \\ \hline
   OTAM   &  ResNet50   &  23.5M  & 7.9G & 2.5H &36.4 \\ 
   CLIP-FSAR   &  ViT/B-32   &  90.2M  & 13.2G &  3.4H & 53.1  \\ 
    \textbf{Ours}	& ViT/B-32 &   18.5M  &   11.9G& 3.0H & 56.5  \\ \hline
\end{tabular}}
\label{training_costs}
\vspace{-14pt}
\normalsize
\end{table}

\subsubsection{\textbf{Zero-shot Performance}}
\label{zero-shot perf}
To investigate the zero-shot performance of the CLIP model, we conduct 5-way zero-shot experiments on the spatial-related datasets (Kinetics, HMDB51, and UCF101)  and the temporal-related dataset (SSv2-Small). Comparative experiments are set up for four mainstream methods, and the results are detailed in Tab.~\ref{zero}. In our experiments, CLIP-Freeze refers to using only the original CLIP model and its pre-trained models. CLIP-FSAR~\cite{wang2023clip}, AIM~\cite{yang2023aim}, and our MA-FSAR only use the text-to-image matcher, which fine-tunes the training sets and performs zero-shot recognition on the testing sets. We find that the original frozen CLIP model demonstrates strong recognition capabilities on spatial-related datasets due to its pre-trained model's robust image background discriminative ability. However, its recognition performance is unsatisfactory with near-random accuracy (20\%) of \textbf{28.8\%} on SSv2-Small, reflecting its lack of capability for temporal modeling. In contrast,  our method MA-FSAR exhibits a notable accuracy improvement of 18.8\%  over the Frozen CLIP model on SSv2-Small, highlighting our approach's powerful temporal relation modeling capability. Additionally, our method also shows some performance improvement on spatial-related datasets. Meanwhile, compared to AIM~\cite{yang2023aim} and CLIP-FSAR~\cite{wang2023clip},  our MA-FSAR also has the best zero-shot performance on various datasets. 

% To demonstrate the significance of applying large-scale foundation pre-trained models in few-shot action recognition, significantly reducing the number of training tasks and dramatically improving recognition accuracy
\subsubsection{\textbf{Comparison of The Number of Training Tasks for Different Methods}}
\label{transfer_nums}
 To demonstrate the rapid convergence of our approach during training in FSAR (\textit{i.e.}, the rapid adaptability to transfer between different tasks), we conduct experiments on SSv2-Small and Kinetics in the 5-way 1-shot task to compare the number of training tasks and accuracy among different methods. Our visual encoder is CLIP-ViT/B-32, and MA-FSAR's temporal alignment metric is OTAM. As shown in Tab.~\ref{Num}, our MA-FSAR achieves at least a 15\% improvement in accuracy compared to other methods that use ImageNet pre-training, while the number of training tasks is only one-fourth of theirs on SSv2-Small. Similarly, on Kinetics, our MA-FSAR achieves at least a 10\% improvement in accuracy while the number of training tasks is only one-tenth of other methods. Based on the above results, applying large-scale foundation models to few-shot recognition is necessary. 

\subsubsection{\textbf{Analysis of Hyperparameters}} \label{Hyperparameters}
We explore the impact of hyperparameters, including scaling factor $r$ (Sec.~\ref{Joint_AD}) and the loss factor $\alpha$ (Sec.~\ref{prediction}). As shown in Tab.~\ref{scale_factor_r},  our MA-FSAR delivers optimal performance across various datasets when the scaling factor $r$ in Joint Adaptation is set to 0.5. As illustrated in Tab.~\ref{loss_factor_a}, we conduct experiments to analyze the impact of the adjustable hyperparameter $\alpha$, which controls metric loss and predictions (Sec.\~ref{prediction}) on representative spatial-related (Kinetics) and temporal-related (SSv2-Small) datasets. The results reveal that the model achieves the highest accuracy when $\alpha$ is set to 0.75 on Kinetics and 0.25 on SSv2-Small.

\subsubsection{\textbf{Comparsion of training costs}} \label{Costs}  We conduct the training costs experiments, including training time (hours) and running GPU memory, among the baseline methods OTAM~\cite{cao2020few}, CLIP-FSAR~\cite{wang2023clip}, and our MA-FSAR in the SSv2-Small 5-way 1-shot task. The training time indicates the time training for 10,000 tasks on a single RTX3090.  As shown in Tab~\ref{training_costs}, our MA-FSAR achieves the best accuracy performance with minimal training costs.

\subsubsection{\textbf{Attention Visualization of MA-FSAR}} \label{Visualization}
Fig.~\ref{fig:vis} shows the attention visualization of our MA-FSAR in the SSv2-Small 5-way 1-shot task. Corresponding to the original RGB images (left), the attention maps of the unimodal full fine-tuning model using CLIP pre-trained weights (middle), which we have mentioned in Sec.~\ref{SM} are compared to the attention maps with our MA-FSAR (right). As illustrated in Figure ~\ref{fig:vis}, the attention maps generated by MA-FSAR prioritize action-related objects and reduce attention to the background and unrelated objects. These observations substantiate the empirical evidence of the efficacy of our MA-FSAR in enhancing semantic and spatiotemporal representation.
\vspace{-10pt}

% by adding lightweight adapters, which can minimize the number of learnable parameters and enable the model to possess the ability to transfer across different tasks quickly. These adapters we design combine information from video-text multimodal sources for temporal and multimodal modeling, each assigned a specific role in learning temporal or multimodal relationships at various positions within the network. To further adapt CLIP for few-shot action recognition, we propose a text-guided prototype construction module that fully leverages CLIP's video-text features to enhance the representation of video prototypes. Our MA-FSAR is designed to be plug-and-play and easily integrated into a wide range of few-shot action recognition temporal alignment metrics. 

\section{Conclusion}
In this work, we propose a novel method, MA-FSAR, to refine CLIP for few-shot action recognition. Our solution, based on the PEFT technique, incorporates a Fine-grained Multimodal Adaptation (FgMA) tailored for FSAR with the enhancement of the action-related temporal and semantic representations, which is fast, efficient, and cost-effective in training. Specifically, we first introduce a Global Temporal Adaptation that processes only the class token to efficiently capture global motion cues. These outputs are fed into the subsequent Local Multimodal Adaptation to guide the local visual tokens in learning spatiotemporal details. Notably, this module can integrate text features specific to the FSAR support set, emphasizing fine-grained semantics associated with actions. At the prototype level, we propose a Text-guided Prototype Construction Module (TPCM)  to further enrich the temporal and semantic characteristics of video prototypes.  Extensive experiments across various task settings and five widely used datasets consistently demonstrate our method's excellent performance in any temporal alignment metric, achieved with minimal trainable parameters.

\noindent\textbf{Limitations.} While our method demonstrates exceptional performance across all datasets, it remains data-driven, relying on fine-tuning a training set to achieve outstanding results on the testing set. Looking ahead, our goal is to explore improved utilization of Large Language Models (LLM) to develop knowledge-driven few-shot action recognition methods, enhancing the model's generalization capabilities.

\vspace{-8pt}

% \clearpage
% \vspace{-5pt}

{\small
	\bibliographystyle{IEEEtran}
	\bibliography{tnnls.bib}
}

\vfill

\end{document}